\documentclass[journal]{IEEEtran}

\usepackage{graphicx}
\usepackage{amsmath}
\usepackage{multirow}
\usepackage{booktabs}
\usepackage{amssymb}
\usepackage{url}
\usepackage{color}
\usepackage{subfigure}
\usepackage{multicol}
\usepackage{xcolor}
\usepackage{algorithm}
\usepackage{algpseudocode}
\usepackage{hyperref}

\ifCLASSINFOpdf

\else

\fi

\begin{document}
	
	\title{Learning to Evaluate Performance of Multi-modal Semantic Localization}
	
	\author{
		Zhiqiang~Yuan,~\IEEEmembership{}
		Wenkai~Zhang,~\IEEEmembership{}
		Chongyang~Li,~\IEEEmembership{}
		Zhaoying~Pan,~\IEEEmembership{}
    	Yongqiang~Mao,~\IEEEmembership{}\\
		Jialiang~Chen,~\IEEEmembership{}
		Shouke~Li,~\IEEEmembership{}
		Hongqi~Wang~\IEEEmembership{Member,~IEEE,}
		and~Xian~Sun,~\IEEEmembership{Senior Member,~IEEE}
		
		\thanks{This work was supported by the National Science Fund for Distinguished Young Scholars under Grant 67125105. \textit{(Corresponding author: Shuoke Li.)}}
		\thanks{Z. Yuan, C. Li, Z. Pan, and Y. Mao are with the Aerospace Information Research Institute, Chinese Academy of Sciences, Beijing 100190, China, the Key Laboratory of Network Information System Technology (NIST), Institute of Electronics, Chinese Academy of Sciences, Beijing 100190, China, University of Chinese Academy of Sciences, Beijing 100190, China and the School of Electronic, Electrical and Communication Engineering, University of Chinese Academy of Sciences, Beijing 100190, China (e-mail: yuanzhiqiang19@mails.ucas.ac.cn; lichongyang99@mail.dlut.edu.cn; panzhaoying17@mails.ucas.ac.cn; maoyongqiang19@mails.ucas.ac.cn.}
		\thanks{W. Zhang, J. Chen, S. Li, H. Wang, and X. Sun are with the Aerospace Information Research Institute, Chinese Academy of Sciences, Beijing 100190, China and the Key Laboratory of Network Information System Technology (NIST), Institute of Electronics, Chinese Academy of Sciences, Beijing 100190, China (e-mail: zhangwk@aircas.ac.cn; chenjl@aircas.ac.cn; lisk@aircas.ac.cn; wiecas@sina.com; sunxian@mail.ie.ac.cn).}
	}
	
	\markboth{}%
	{Shell \MakeLowercase{\textit{et al.}}: Bare Demo of IEEEtran.cls for IEEE Journals}
	
	\maketitle
	
	\begin{abstract}
		Semantic localization (SeLo) refers to the task of obtaining the most relevant locations in large-scale remote sensing (RS) images using semantic information such as text.
		As an emerging task based on cross-modal retrieval, SeLo achieves semantic-level retrieval with only caption-level annotation, which demonstrates its great potential in unifying downstream tasks.
		\textcolor{black}{Although SeLo has been carried out successively, but there is currently no work has systematically explores and analyzes this urgent direction.}		
		In this paper, we thoroughly study this field and provide a complete benchmark in terms of metrics and testdata to advance the SeLo task.
		Firstly, based on the characteristics of this task, we propose multiple discriminative evaluation metrics to quantify the performance of the SeLo task.
		The devised significant area proportion, attention shift distance, and discrete attention distance are utilized to evaluate the generated SeLo map from pixel-level and region-level.
		Next, to provide standard evaluation data for the SeLo task, we contribute a diverse, multi-semantic, multi-objective Semantic Localization Testset (AIR-SLT).
		AIR-SLT consists of 22 large-scale RS images and 59 test cases with different semantics, which aims to provide a comprehensive evaluations for retrieval models.
		Finally, we analyze the SeLo performance of RS cross-modal retrieval models in detail, explore the impact of different variables on this task, and provide a complete benchmark for the SeLo task.
		\textcolor{black}{We have also established a new paradigm for RS referring expression comprehension, and demonstrated the great advantage of SeLo in semantics through combining it with tasks such as detection and road extraction.}
		The proposed evaluation metrics, semantic localization testsets, and corresponding scripts have been open to access at \textcolor{black}{https://github.com/xiaoyuan1996/SemanticLocalizationMetrics}.

	\end{abstract}
	

	\IEEEpeerreviewmaketitle
	
	\section{Introduction}

	\begin{figure}[!t]
		\centering
		\includegraphics [width=3.5in]{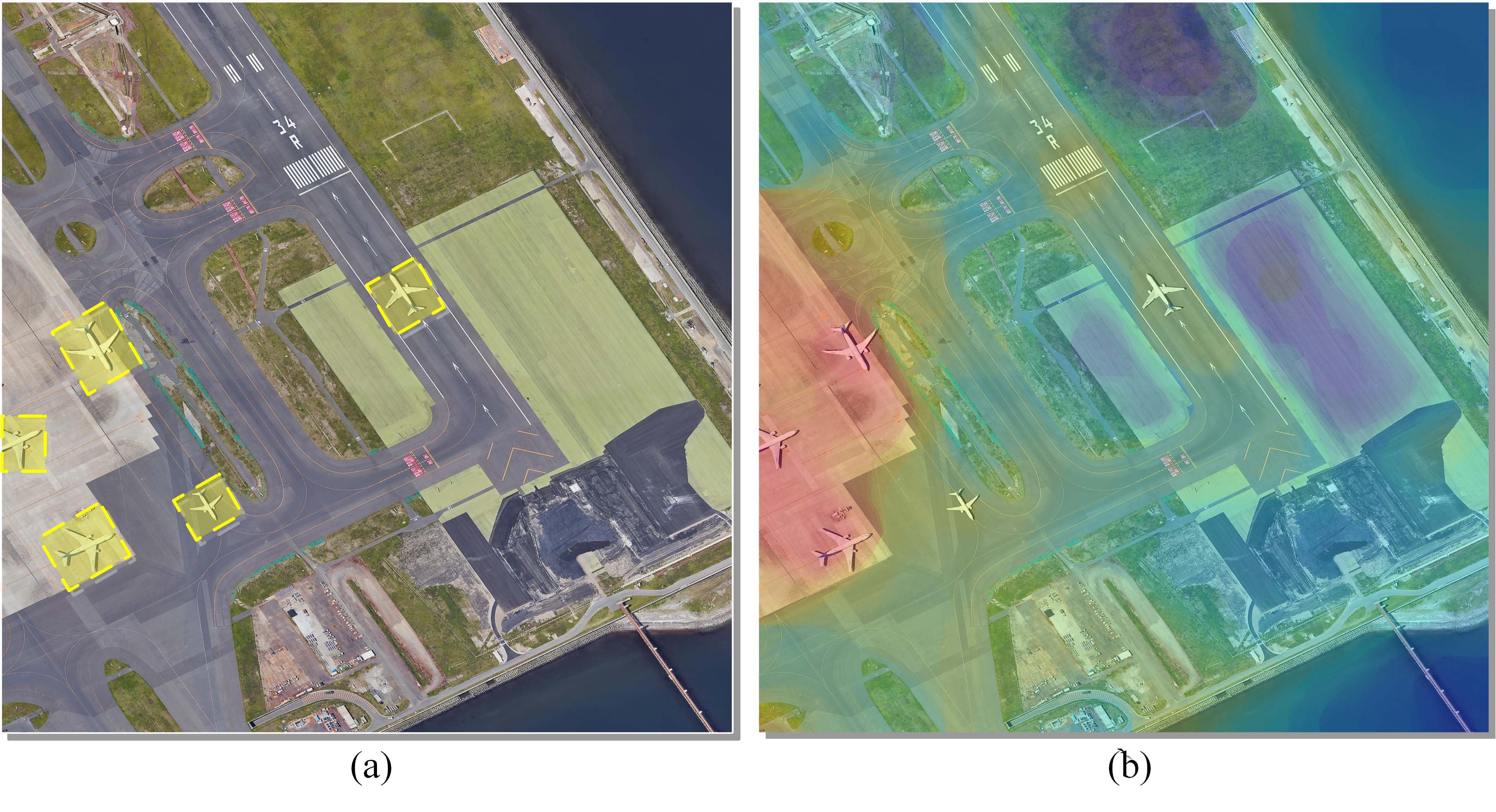}
		\caption{
			(a) Results of airplane detection.
			(b) Results of semantic localization with query of ``white planes parked in the open space of the white airport''.
			Compared with tasks such as detection, SeLo achieves semantic-level retrieval with only caption-level annotation during training, which can adapt to higher-level retrieval tasks.
		}
		\label{compare}
	\end{figure}
	
	In recent years, remote sensing (RS) images are becoming more and more accessible, which greatly improves people's perception of the earth \cite{Remote sensing big}\cite{Big data for}.
	In order to obtain the valuable objects from numerous data, RS image retrieval (RSIR) has become a hotspot in recent years \cite{Image retrieval from}.
	As a branch of RSIR, RS cross-modal text-image retrieval (RSCTIR) has gradually become an emerging direction of research due to its superiority in human-computer interaction \cite{A Novel SVM-Based}\cite{Automatic Feature Extraction}.

	Leading RSCTIR methods currently can be divided into caption-based and embedding-based methods \cite{Unsupervised Contrastive Hashing}.
	Caption-based RSCTIR methods generally apply caption generators to describe RS images, and obtain the retrieval results by calculating the BLEU \cite{Bleu} score between the query text and the generated captions.
	Considering the correlation between ground elements at different scales, Shi $et\ al.$ \cite{Can a machine} proposed a caption generation framework to generate robust RS captions.
	In order to make the generated caption interpretable, Wang $et\ al.$ \cite{Word-Sentence Framework for} designed an explainable word-sentence framework, decomposing the task into a word classification task and a word sorting task.
	Embedding-based methods embed the RS images into a learnable representation space to find the nearest neighbors with the retrieved text.
	To fully explore the latent correspondence between RS images and text, Cheng $et\ al.$ \cite{A Deep Semantic} proposed a semantic alignment module to filter and optimize features so that to obtain higher retrieval accuracy.
	Further on the fine-grained retrieval, researchers \cite{Exploring a fine-grained} proposed an asymmetric multi-scale feature matching network to calculate the distance of multi-modal embedding vectors.
	\textcolor{black}{Compared with caption-based methods, embedding-based methods as a one-stage retrieval method greatly preserve the source features, which makes it the preferred way for RSCTIR in recent two years.}

	On the basis of the RSCTIR task, the semantic localization (SeLo) task has recently been proposed while gaining considerable attention.
	The SeLo task is defined as using cross-modal information such as text to locate semantically similar regions in large-scale RS scenes.
	\textcolor{black}{This task is usually based on RS cross-modal retrieval (RSCR) and thus is regarded as a higher-level retrieval task than RSCR.}
	Yuan $et\ al.$ \cite{Exploring a fine-grained} was the first to apply SeLo to detection tasks, and obtained a considerable retrieval effect.
	To further reduce the inference complexity in the SeLo task, Yuan $et\ al.$ \cite{A Lightweight Multi-Scale} proposed a lightweight RSCTIR method which greatly improves the retrieval speed.
	As shown in Fig. \ref{compare}, even if SeLo only uses caption-level annotations during training, it enables interesting and exciting semantic-level retrieval compared to detection and other scene-specific retrieval tasks.
	Given the great advantage of SeLo in semantic retrieval, this task has become an urgent research hospot in RS field.

	Vanilla exploration has been carried out on the SeLo task, but the current research is still limited to qualitative analysis \cite{Remote Sensing Cross-Modal}. 
	In previous work, researchers qualitatively analyzed the recall metrics of the retrieval models and compared SeLo performance only from a visual perspective.
	Therefore, the lack of quantitative analysis makes the SeLo task impossible to decouple from retrieval, which greatly hinders the development of this task.
	In such cases, it is particularly important to conduct a reasonable analysis of the task and propose a set of discriminative evaluation metrics to quantify the performance of the SeLo task.
	To this end, we firstly systematically analyze and study SeLo and explore the advantages and disadvantages of this task in detail.
	Next, the novel significant area proportion, attention shift distance and discrete attention distance are proposed to comprehensively quantify the performance of SeLo task.
	To the best of our knowledge, this paper is the first integral work that completely focuses on the emerging SeLo task.

	Existing SeLo work is not convincing enough in qualitative evaluation, which is caused by the lack of well-established SeLo testsets in the RS field.
	Aiming for a unified and comprehensive test for this task, we contribute a multi-semantic, multi-scene, multi-objective Semantic Localization Testset (AIR-SLT).
	AIR-SLT contains 22 large-scale RS images and provides 59 test cases with a total of 103 semantic-level bounding boxes to evaluate the SeLo task from different perspectives and in different scenarios.
	Different from traditional detection datasets, the proposed AIR-SLT starts with the target relationship and thoroughly evaluates the relationship understanding ability of the retrieval model.
	Currently, AIR-SLT is the only multi-modal large-scale semantic localization testset with crucial relational attributes in RS.
		
	Justifiable experiments are further performed to provide numerous baselines, which greatly promote the development of the task.
	We explore the effects of different segmentation scales, different trainsets, and different retrieval models on the performance of SeLo, and exhaustively mine the time consume of the task at various stages.
	We also attempt to combine SeLo with other tasks such as detection, and demonstrate that SeLo maps with semantic information can provide semantic priors for these tasks.
	Furthermore, all relevant materials about SeLo are contributed and have been open to access to advance this task.
	We hope that more researchers can focus on this field, which may be a good opportunity to achieve multi-task unification under \textcolor{black}{enormous RS text-image data and pre-trained RSCR models}.
	
	In general, the main contributions of our work are as follows:
	\begin{itemize}
		
		\item We systematically model and study semantic localization in detail, and propose multiple discriminative evaluation metrics to quantify this task based on significant area proportion, attention shift distance, and discrete attention distance.
		
		\item We provide a multi-semantic, multi-scene, and multi-objective semantic localization testset (AIR-SLT), thus enabling comprehensive evaluation for cross-modal retrieval models from different perspectives and in different scenarios.
		
		\item To the best of our knowledge, we are the first to fully focus on the SeLo task and provide sufficient data support and complete variable analysis for this field.
		
	\end{itemize}

	Greatness experiments have been carried out to provide massive benchmarks for further research.
	The rest of the paper is organized as follows:
	Section II briefly summarizes related works involved with SeLo.
	In Section III, the formulation of SeLo and the corresponding metrics are introduced.
	Next, in Section IV, we provide exhaustive information of the contributed AIR-SLT. 
	Furthermore, we analyze the SeLo task in detail with extensive experiments in Section V and provide a large amount of data for this task from qualitative and quantitative.
	At last, the conclusions are given in Section VI.

	\section{Related Work}
	In this section, we first review the previous research on RS cross-modal retrieval in three aspects \cite{Exploiting deep features}: RS image-image retrieval, RS audio-image retrieval, and RS text-image retrieval. 
    \textcolor{black}{Next, we summarize the current progress in semantic localization.}

	\subsection{\textcolor{black}{Remote Sensing Cross-Modal Retrieval}}
	\textbf{RS Cross-modal Image-image Retrieval:}
	RS cross-modal image-image retrieval (RSCIIR) refers to retrieve RS images by using images in other modalities, such as using synthetic aperture radar images to retrieve RS image in natural scenes.
	In prior work \cite{Learning source-invariant deep}, Li $et\ al.$ designed a new convolutional neural network (CNN) \cite{Understanding of a} based on source-invariant hash method and retrieved cross-modal RS image on dual-source datasets.
	Zhang $et\ al.$ \cite{Hyperspectral remote sensing} and Zhou $et\ al.$ \cite{Learning low dimensional} exploited the image content and low-level features to retrieve RS images.
	Xiong $et\ al.$ \cite{A deep crossmodality} converted three channels of optical images into four different types of single-channel images and combined triplet loss with hash functions to improve the retrieval efficiency.
	Later, Demir and Bruzzone \cite{Hashing-based scalable remote} introduced hashing-based approximate nearest neighbor search to retrieve RS images efficiently and accurately.
	In \cite{Learning to translate}, Xiong $et\ al.$ proposed a cycle identity generation adversarial network to reduce data drift during multi-source RS image retrieval.
	Even though the methods for RSCIIR task are quite mature, this task is still limited by the low-level features acquisition \cite{Image retrieval from}.
	Therefore, the RSCIIR methods may not be able to perform retrieval efficiently in specific retrieval scenarios even if it obtains great retrieval results.

	\textbf{RS Cross-modal Audio-image Retrieval:}
	Considering the audio information for RS image retrieval, Mao $et\ al.$ \cite{Deep cross-modal retrieval} contributed a large-scale hand-annotated RS voice-image dataset for cross-modal RS audio-image retrieval (RSCAIR).
	Chen $et\ al.$ \cite{A deep hashing} proposed a deep triplet-based hashing to integrate hash code learning and relative relationship learning into the end-to-end network.
	Furthermore, Guo $et\ al.$ \cite{Jointly learning of} designed an audio-image similarity calculation model and then applied a neural network for fusion and classification.
	Regarding the problem of insufficient utilization of cross-modal similarity, Chen $et\ al.$ \cite{Deep cross-modal image-voice} solved this problem by exploiting the multi-scale context information.
	Although the RSCAIR methods bring excellent convenience to users, a certain amount of noise may be generated along with the direct inputs of audio signal.
	
	\textbf{RS Cross-modal Text-image Retrieval:}
	RS Cross-modal Text Image Retrieval (RSCTIR) methods can be further divided into caption-based methods and embedding-based methods.
	Caption-based methods generates a caption for each RS image automatically, and then rank the text similarity between the query and the captions during retrieval\cite{Natural language description}\cite{Multi-scale cropping mechanism}\cite{Semantic descriptions of}.
	Shi and Zou \cite{Can a machine} designed a RS image caption framework by modeling the interaction of RS image attributes.
	Lu $et\ al.$ \cite{Exploring models and} proposed the largest dataset RSICD for RS image caption.
	Hoxha $et\ al.$ \cite{A new CNN-RNN} applied a CNN-RNN framework that combined with beam search to generate multiple captions, and then selected the best caption by utilizing prior similarity.
	Li $et\ al.$ \cite{Truncation cross entropy} constructed a truncation cross entropy (TCE) loss to alleviate the overfitting problem.
	Lu $et\ al.$ \cite{Sound active attention} established a sound active attention framework for more specific caption generation.
	Although caption-based RSCTIR is relatively mature, the fine-grainedness of the generated sentences cannot cover the entire image, which causes inevitable information loss during modality transitions.

	The embedded-based RSCTIR methods refer to mapping RS image and text into the same high-dimensional space and measuring the cross-modal similarity by appropriate distance.
	Abdullah $et\ al.$ \cite{TextRS} applied deep bidirectional triplet network to calculate the similarity of text and image in RS scene.
	In order to further explore the potential correspondence between RS images and text, Cheng $et\ al.$ \cite{A Deep Semantic} proposed a semantic alignment module to get a more discriminative feature representation.
    Yuan $et\ al.$ \cite{A Lightweight Multi-Scale} proposed a lightweight text-image retrieval model, which realized fast RS cross-modal retrieval, and improved the retrieval performance by using knowledge extraction and contrast learning. 
    Further, Yuan $et\ al.$ \cite{Remote Sensing Cross-Modal} added the denoised detection information to the RS image representation, which greatly improved the retrieval accuracy.
    The single-stage calculation of embedded-based RSCTIR greatly reduces the loss of information transformation and becomes the main cross-modal retrieval method in recent years. 

	\subsection{Multi-Modal Semantic Localization}
	
	Semantic localization refers to the task of obtaining the most relevant locations in large-scale RS images using semantic information such as text.
	\textcolor{black}{Unlike object detection and segmentation tasks with pixel-level orientation \cite{Hyperspectral Target Detection}\cite{Historical Information-Guided Class-Incremental}, SeLo aims to retrieve relevant regions from a semantic level.}
	\textcolor{black}{Referring expression understanding (REC) in natural scenes also attempts object retrieval from a semantic perspective by multimodal representation \cite{Uniter}\cite{Fusion of Detected}, but SeLo has much less supervised information than REC, which is the reason for considering SeLo as a semi-supervised REC task.}
	After multi-scale cropping of large RS images, Yuan $et\ al.$ \cite{Exploring a fine-grained} proposed a framework to generate SeLo map based on cross-modal retrieval, which is the earliest implementation of semantic localization.
	Next, to improve the inference speed of SeLo tasks, Yuan $et\ al.$ \cite{A Lightweight Multi-Scale} designed a lightweight RSCTIR model, and improved the model from the perspective of knowledge distillation and negative sampling.
	\textcolor{black}{In \cite{Remote Sensing Cross-Modal}, the authors qualitatively compare the visualization results of the method and others, thereby verifying the effect of the proposed fusion module.}
	Although RS multimodal semantic localization is a recently emerging task, \textcolor{black}{above works only judge the task from a qualitative analysis perspective}, which lacks discriminative quantitative metrics and unified baselines.
	Accordingly, it is urgent and important to develop a set of quantitative evaluation indicators for this task.
    For this purpose, we establish an blameless evaluation system to quantify the performance of Selo task, and conduct a systematic and detailed research and analysis of this task.

	\begin{figure*}[!t]
		\centering
		\includegraphics [width=6.8in]{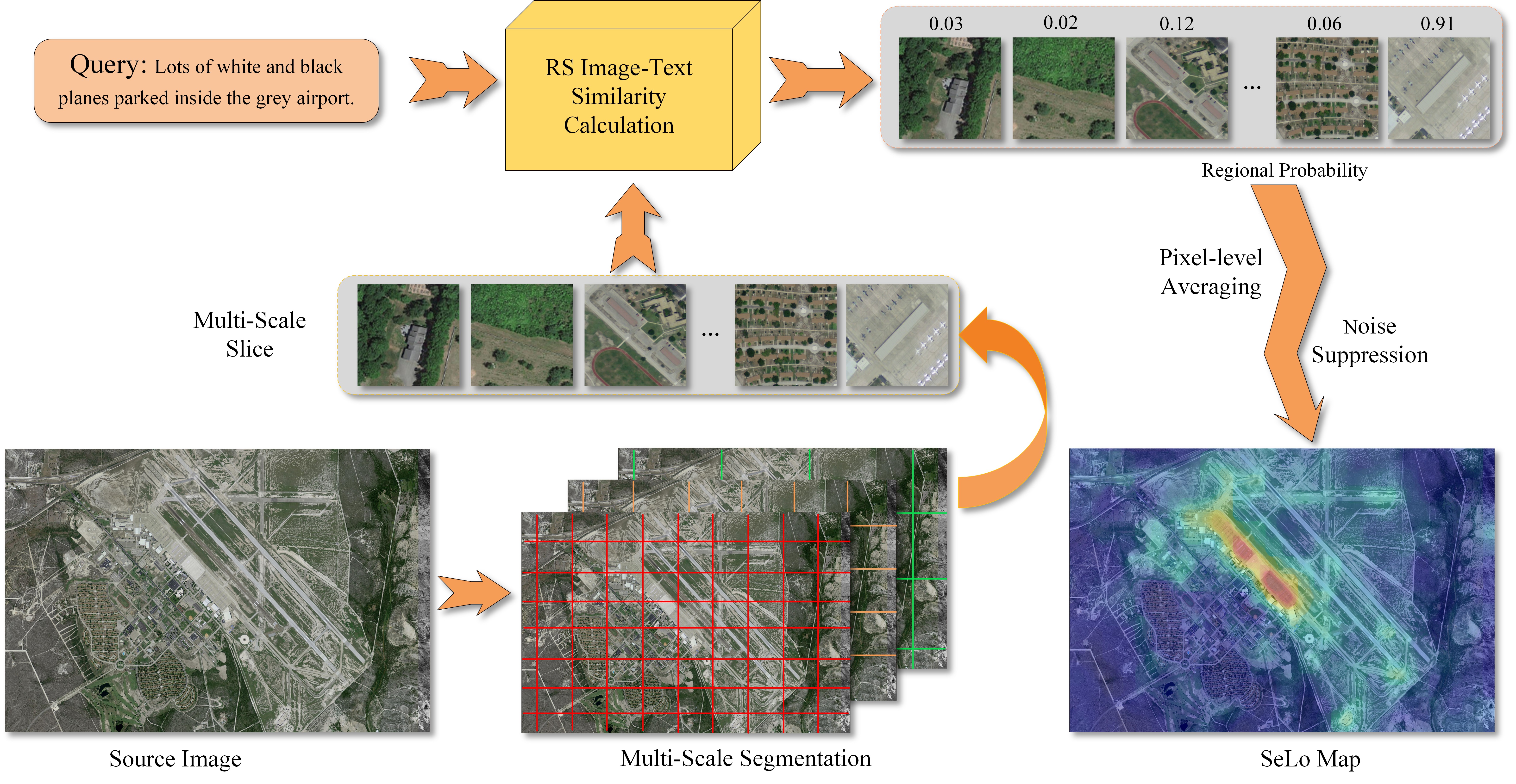}
		\caption{
			Framework of Semantic Localization.
			After multi-scale segmentation of large RS images, we perform cross-modal similarity calculation on query and multiple slices.
			The calculated regional probabilities are then utilized to pixel-level averaging, which generates the SeLo map after further noise suppression.
		}
		\label{SeLo}
	\end{figure*}

	\section{Discriminative Evaluation Methods}\label{sec-3}
	This section provides multiple discriminative quantitative metric from pixel-level and region-level for semantic localization task.
	We introduce the proposed evaluation metrics from:
	A) Formulation of SeLo;
	B) Improving RoI with significant area proportion;
	C) Reducing attention shift distance;
	D) Minimizing the discrete attention distance;
	and E) Indicator integration.
			
	\subsection{Formulation of Semantic Localization}
	
	SeLo task refers to the task of obtaining the most relevant locations in large-scale RS images by specific cross-modal query.
	\textcolor{black}{Yuan $et\ al.$ \cite{Exploring a fine-grained} first applied cross-modal retrieval to accomplish this task, which verified the feasibility of the SeLo performance on sub-tasks such as object detection and road extraction.}
	Generally, SeLo task is based on cross-modal retrieval, and only the caption-level annotations are required during training to achieve downstream tasks.
	\textcolor{black}{Compared to specific retrieval tasks such as detection \cite{ A General Gaussian}, SeLo realizes semantic-level retrieval as shown in Fig. \ref{SeLo}, which is considered to be a higher-level retrieval task with promising development prospects.}
	\textcolor{black}{Since the SeLo task is currently built on the basis of RSCR}, we first model the RSCR task and then expand it to the SeLo.
	In this paper, we take RSCTIR as an example for cross-modal retrieval.

	\begin{figure}[!t]
		\centering
		\includegraphics [width=3.2in]{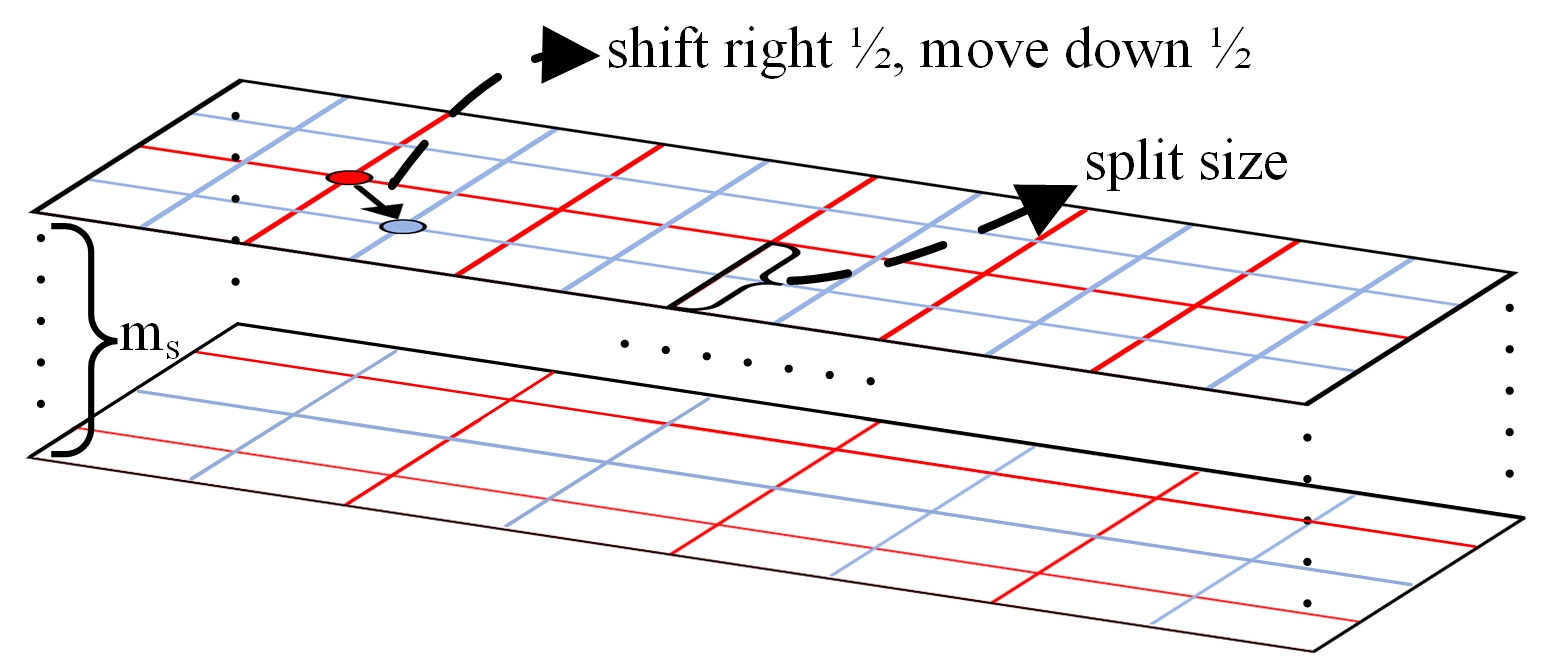}
		\caption{
			Multi-scale segmentation.
			After the first cropping, we move the vertices of the cropping box by 1/2 grid and regard the above operation as a round.
			Then we change the size of the slice and crop it with multiple scales.
		}
		\label{seg}
	\end{figure}

	
	To calculate the similarity between images and texts, RSCTIR models are successively proposed to provide suitable multi-modal embeddings.
	Firstly, given an slice $I_s$ and retrieved text $T_q$, \textcolor{black}{we map them to the representation space $\mathbb{R}^{d}$}, which is represented as:
	\begin{equation}
	f^{s}_{v} = W_v I_s, f_t = W_t T_q, (f^{s}_{v},f_t \in \mathbb{R}^{d})
	\end{equation}
	where $W_v$ and $W_t$ is constructed to project $I_s$ and $T_q$ to the embedded feature space.
	Then, the element-wise product is used to calculate the cross-modal similarity $S_s$ of the two embedding vectors:
	\begin{equation}
	S_s = l_2(f_{v}^{s}) \odot l_2(f_t)
	\end{equation}
	where $\odot$ denotes the element-wise product of matrices, and $l_2(x)$ represents performing $L2$ norm \cite{Analysis of Local} on feature $x$.
	
	\begin{figure*}[!t]
		\centering
		\includegraphics [width=6.8in]{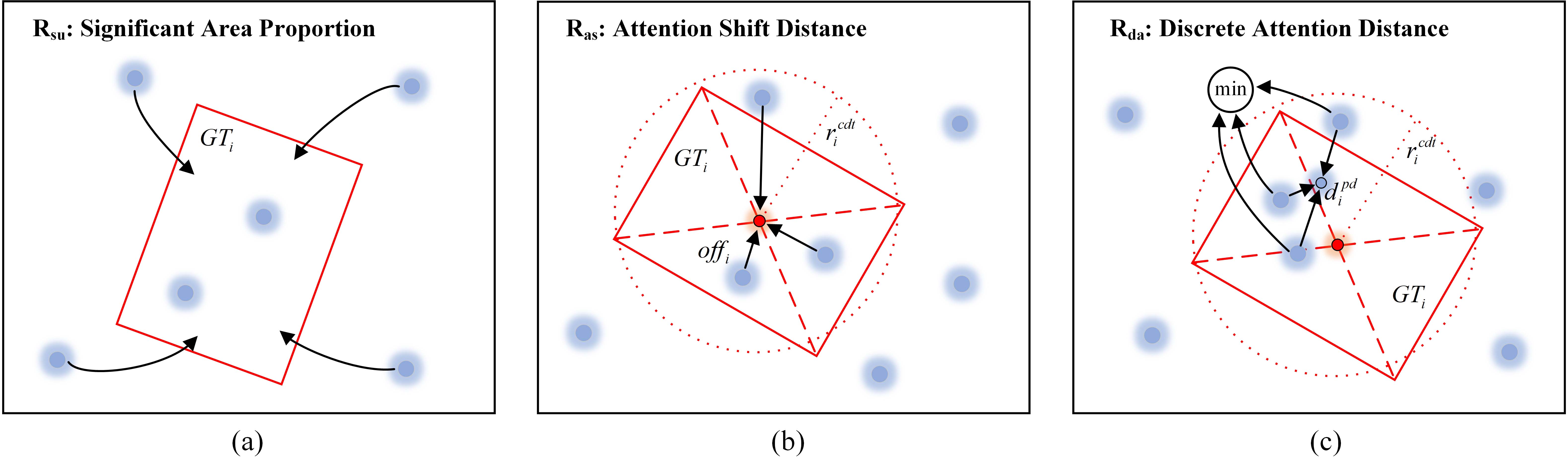}
		\caption{
			Three proposed evaluation metrics for semantic localization.
			(a) $R_{su}$ aims to calculate the attention ratio of the ground-truth area to the non-GT area.
			(b) $R_{as}$ attempts to quantify the shift distance of the attention from the GT center.
			(c) $R_{da}$ evaluates the discreteness of the generated attention from probability divergence distance and candidate attention number.
		}
		\label{indicators}
	\end{figure*}

	Further, to obtain SeLo map, we perform multi-scale segmentation of large RS images and conduct multiple cross-modal similarity calculations.
	As shown in Fig. \ref{seg}, for a large image $I$, we crop it with different scales and offsets to get $N_s$ slices:
	\begin{equation}
	\{I_s\}_{N_s} = SEG_{s_i, off_j}(I) 
	\end{equation}
	\textcolor{black}{where $s_i \in [s_1, ..., s_m]$ represents the slice scale, which we discuss further in the appendix A.
	$off_j \in [off_1, ..., off_m]$ represents the slice start offset, and $SEG$ is used to denote multi-scale and multi-offset cropping.} 
	We calculate the similarity of $N_s$ slices with the retrieval text $T$ respectively to obtain the probability distribution $\{S_s\}_{N_s}$ of each slice.
	After pixel-level averaging of multiple similarity, median filtering is utilized to filter shock signals in the probability map, providing a smoother probability distribution.
	\textcolor{black}{Then we normalize the probability distribution to get the SeLo map $SM$, and the above process is denoted as:
	\begin{equation}
	SM =  \mathbb{N}(\partial ( \Gamma^{N_s}_1 (S_s)  ) )
	\end{equation}
	}where $\Gamma^n_1(x)$ represents the pixel-level average of $n$ regional similarity, $\partial(x)$ means smoothing the probability map by median filter, and $\mathbb{N}(x)$ means to normalize the probability map $x$.

	Compared with tasks such as object detection and semantic segmentation, the generated SeLo map contains relational attributes, so that objects and relationships under specific scenes can be retrieved based on semantic information.
	While compared with tasks such as referring expressions comprehension in natural scenes \cite{MAttNet}\cite{A Real-Time Global}, SeLo task only utilizes caption-level annotations, and the former further uses additional information such as bounding boxes.
	In summary, the SeLo can be regarded as a weakly supervised retrieval task with higher-level semantics, which has great application potential in object retrieval.
	Although studies such as \cite{Exploring a fine-grained}\cite{A Lightweight Multi-Scale}\cite{Remote Sensing Cross-Modal} have explored this field qualitatively, there is still no unified framework and evaluation basis for this task so far.
	To provide a quantitative evaluation for SeLo, we construct diverse and reasonable indicators as shown in Fig. \ref{indicators}, establish a complete Semantic Localization Testset and provide extensive data to promote the development of this task.
	

	\subsection{Significant Area Proportion}
	To evaluate the SeLo task, we first attempt to measure the SeLo map at the pixel level.
	Suppose there are $N$ ground-truth (GT) boxes in a RS image, which is denoted as $GT_1, ..., GT_i, ..., GT_N$, and each of the GT boxes contains $M$ nodes.
	In order to make the model focus all attention on the ground-truth regions, a basic idea is to maximize the ratio of attention in the GT regions to the non-GT regions.
	Considering the great difference of size between the RS images, it is necessary to set a norm term to offset the difference in the number of pixels on the evaluation results.
	
	Considering on the above factors, we define the ratio of significant areas to non-GT areas as the significant ratio $R_{su}$.
	Specifically, we define $R_{su}$ as:
	\begin{equation}
	t_l =  \frac{ \sum_{i=1}^{N} \oint_{GT_{i}} v}{\oint_{all} {v}-\sum_{i=1}^{N}\oint_{GT_{i}} v + eps}
	\end{equation}
	\begin{equation}
	t_r =  \frac{HW-\sum_{i=1}^{N}\oint_{GT_{i}} 1}{\sum_{i=1}^{N}\oint_{GT_{i}} 1}
	\end{equation}
	\begin{equation}
	R_{su} = - exp( - \alpha \cdot t_l \cdot t_r ) + 1
	\end{equation}
	where $\oint_{GT_{i}}v$ is defined as the sum of all probability in the GT boxes of $GT_i$, and $\oint_{all}v$ is the total probability of the entire SeLo map.
	The $t_l$ term calculates the ratio of the probability sum of the GT boxes to the non-GT regions, $eps$ is the floating-point relative error limit to avoid the division by zero.
	In order to offset the difference of image size to the final indicator, we introduce size information in the $t_r$ term to stabilize the indicator $R_{su}$ facing different sizes, where $\oint_{GT_{i}} 1$ represents the total number of pixels in the GT box $GT_i$.
	Finally, since the upper and lower bounds in the exponential term are zero and infinity, we use the exponential function to convert it to a range of 0-1 to get a more compact indicator representation, and the parameter $\alpha$ is utilized to set the critical threshold for positive and negative examples.
	
	The significant area proportion $R_{su}$ aims to calculate the probability ratio of the GT region and the other region.
	$R_{su}$ tends to 1 when the model tries to put all attention on the GT region, and 0 otherwise.
	\textcolor{black}{Note that Rsu is related to both the GT region and the non-GT region, which analyzes the dynamic changes of both.
	This indicator starts from the pixel-level and directly reflects the distribution of the generated attention, which is a significantly important indicator for SeLo task.}
	In our opinion, high $R_{su}$ is necessary but not sufficient for well SeLo map.
	When there are multiple GT boxes in the RS image, and only one GT box with non-zero probability, the indicators at this time may not be able to objectively evaluate the SeLo task.
	But this case is a special case, and later we introduce the indicator $R_{as}$ to offset the inexact evaluation in this case, so $R_{su}$ is still a reliable SeLo indicator.
	
	\subsection{Attention Shift Distance}
	Unlike segmentation and detection, which have clear localization targets, the retrieval target of SeLo is not clear enough.
	The semantic activation centers derived from the model may be shifted from the human prediction, but usually this errors is within acceptable limits.
	If the SeLo task is only evaluated from the attention ratio of the GT area, it will be impossible to measure the shift of attention.
	Therefore, we attempt to quantify the distance between the center of the $GT_i$ and the proximate top-k attentions.
	When this distance is small enough, the attention generated by the model will focus on the corresponding GT areas.
	
	Motivated by this, we define the above existing distance as the attention shift distance, which is denoted as $R_{as}$ to measure the degree of attention shift.
	Specifically, for the GT box $GT_i$, we first calculate its center coordinates ($c_i^x, c_i^y$), which is defined as:
	\begin{equation}
	c_i^x = \frac{1}{M} \sum^M_{m=0} {p_{i, m}^x}, c_i^y = \frac{1}{M} \sum^M_{m=0} {p_{i, m}^y}
	\end{equation}
	where $c_i^x$ is the abscissa of the $m^{th}$ node in $GT_i$, and $c_i^y$ is the ordinate of the corresponding $m^{th}$ node.
	Next, we calculate attention regions within a certain range from the GT center ($c_i^x, c_i^y$).
	If the circumcircle of the GT region is directly leveraged as the candidate region, the range to be calculated for regions with larger aspect ratios will be extremely large, which is unfair for GT boxes with smaller aspect ratios.
	To this end, we define the candidate region radius of $GT_i$ as:
	\begin{equation}
	r^{cdt}_i = \frac{\Im}{M} \sum^{M}_{m=0} {\| (p_{i, m}^x, p_{i, m}^y), (c_i^x, c_i^y)   \|_2}
	\end{equation}
	where $\| p_a, p_b \|_2$ is the Euclidean distance of the point $p_a$ and $p_b$,
	$\Im$ is defined as the expansion factor of the candidate region, which is utilized to scale the average radius.
	After the center and candidate radius are obtained, the corresponding attention candidate region can be calculated.

	Further, we locate the attention centers in the candidate regions of the $GT_i$, so as to calculate the attention shift distance.
	In order to find the attention center, we need to find the local maxima in the probability map.
	At the same time, to avoid the influence of small probability local maxima on the final result, the probability threshold $\varrho$ needs to be set to filter the results to obtain $K$ probability centers:
	\begin{equation}
	\{att^x_{i,k}, att^y_{i,k}\}_{k=1}^{K} = \Upsilon_{ p  > \varrho}( \psi(GT_i)), p_j \in GT_i
	\end{equation}
	where $\psi(GT_i)$ represents all local maxima in the $GT_i$, and $\Upsilon_{ p  > \varrho}(p_j)$ is defined as a filter, which means that the probability center is discarded when $p_j$ is lower than the probability threshold $\varrho$.
	$att_{i,k}$ is defined as the $k^{th}$ center of attention within the candidate region of $GT_i$.
	Next, we calculate the average offset distance $off_i$ between the GT center and the $K$ probability centers:
	\begin{equation}
	off_i = \sum^K_{k=1}\frac{\|(att^x_{i,k}, att^y_{i,k}), (c_i^x, c_i^y)\|_2} { r_i^{dct}}
	\end{equation}
	$off_i$ considers the average offset distance between each probability center of the candidate region and the GT center.
	When the distance is smaller, the attention of the model is more focused.
	
	However, the current $off_i$ is linear, and for the SeLo task, a larger $off_i$ indicates worse performance, but a small $off_i$ is acceptable due to the target ambiguity on the SeLo task.
	Based on this consideration, we nonlinearize the average shift distance and merge the $off_i$ of multiple GT regions to obtain the attention shift distance $R_{as}$ :
	\begin{equation}\label{eq12}
	R_{as}=\frac{1}{N}\sum_{i=1}^N\frac{e^{off_i \ \beta}-1}{e^\beta-1}
	\end{equation}
	where $\beta$ is the nonlinear coefficient.
	
	$R_{as}$ aims to quantify the deviation of the probability center of the SeLo map from the GT center.
	$R_{as}$ is close to 0 when the model places the attention as close to the GT center as possible, and close to 1 otherwise.
	\textcolor{black}{In the case of GT center is the background instead of the object, such as a query with a relatively vague entity, the GT center at this time is indeed relatively vague.
	Attention may deviate from the GT center in such scenarios, however, the actual localization performance is insensitive to such mismatch.
	At this time, nonlinearize operation (\ref{eq12}) keeps the stability when the attention is within a certain range from the GT center.
	When the attention is far from the GT center, $R_{as}$ has a large value and gradient.
	When the attention is close to the GT center, $R_{as}$ has a small value, and the gradient of $R_{as}$ at this time is also small.}
	The attention of a well SeLo map is focused on the GT center, which is positively correlated with $R_{as}$.

	\subsection{Discrete Attention Distance}
	When performing the SeLo task, the attention computed by the model may be divergent in two cases.
	First, this situation exists when the model is randomly initialized, which will have multiple attention centers in the SeLo map with a random distribution.
	In addition, when the target relationship described in query does not exist in the large RS image, the min-max difference in the pixel-level similarity is small, and divergence may also occur after normalization.
	Although the introduced metrics can measure SeLo from both the pixel-level and attention shift, these metrics cannot be used to evaluate when the attention is locally divergent.
	A good SeLo map must focus attention, which means a single attention is concentrated on the desired target, rather than multiple scattered attention distributed around the target.
	To this end, we introduce the discrete attention distance $R_{da}$ to quantify this situation.
	
	For the extreme case where the number of attention $\ell$ in the candidate region is 0, which means that there is no attention in this region, we set $R_{da}$ as the minimum value of 0.
	Since the previous two indicators aim to force the center of attention to move closer to the center of the GT areas, the indicator $R_{da}$ only aims to make the number of attention of candidate regions to 1, in which case we take $\ell$ as the maximum value of 1.
	When $\ell > 2$, the degree of attention dispersion can be measured in two aspects: the focused number in the candidate region of $GT_i$ and the deviation between these attentions.
	
	\begin{figure*}[!t]
	\centering
	\includegraphics [width=6.8in]{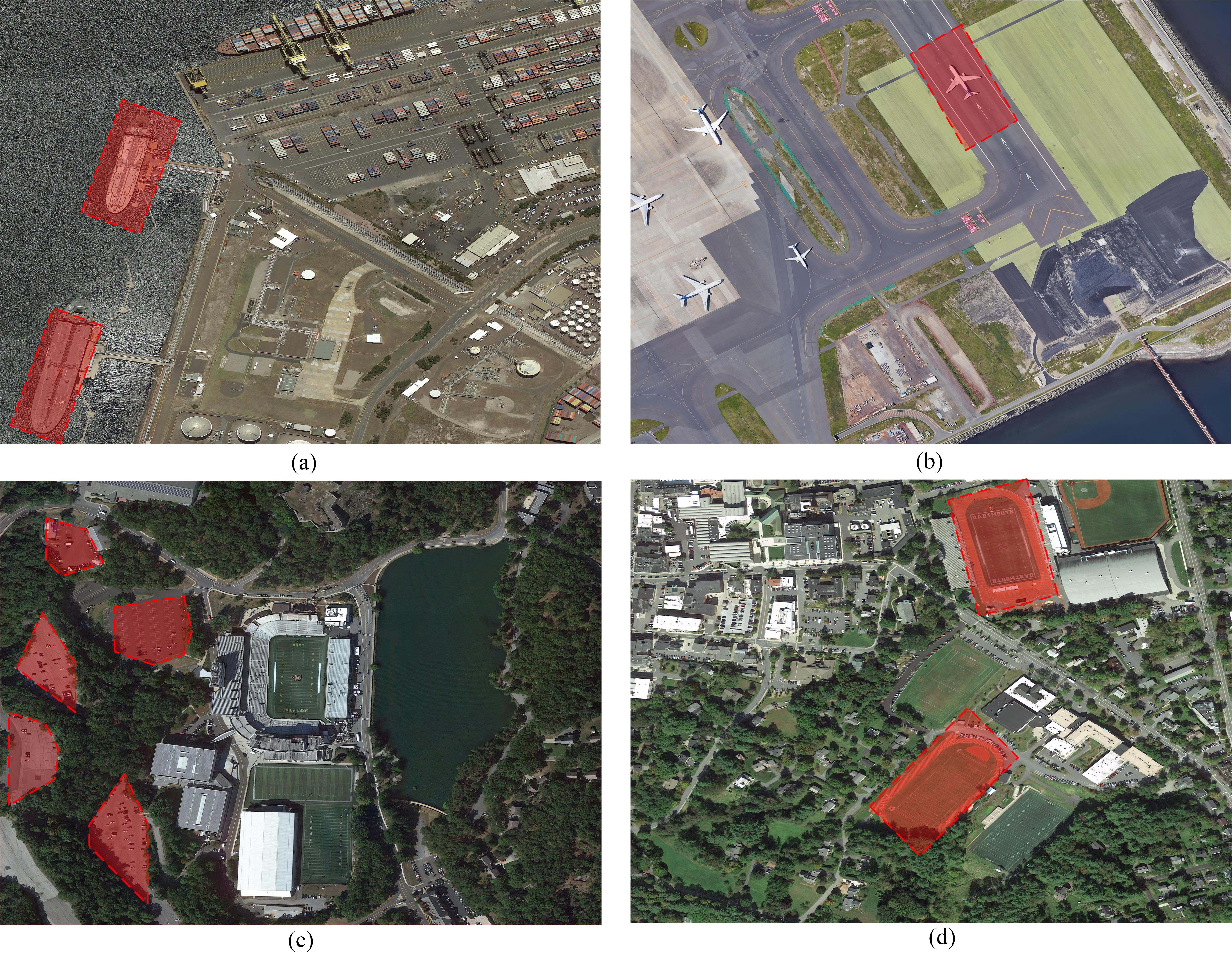}
	\caption{
		Four samples of Semantic Localization Testset.
		(a) Query: ``ships without cargo floating on the black sea are docked in the port''.
		(b) Query: ``a white airplane ready to take off on a grayblack runway''.
		(c) Query: ``some cars are parked in a parking lot surrounded by green woods''.   
		(d) Query: ``the green football field is surrounded by a red track''.        
	}
	\label{sample}
\end{figure*}

	Specifically, after obtaining the $K$ probability centers within the candidate radius $r_i^{cdt}$, we first calculate the cluster centers $(\widetilde{c^x_i}, \widetilde{c^y_i}$) of these attentions:
	\begin{equation}
	\widetilde{c^x_i} = \frac{1}{K} \sum^K_{k=0} {att_{i, k}^x}, \widetilde{c^y_i} = \frac{1}{K} \sum^K_{k=0} {att_{i, k}^y}
	\end{equation}
	Next, we consider the distance of each attention center from the cluster center, and divide it by $r_i^{cdt}$ to convert this distance into the divergence degree of a single attention.
	The probability divergence distance $d_i^{pd}$ in the $GT_i$ region can be obtained after averaging the divergence degree, which is represented as:
	\begin{equation}
	d^{pd}_i = \frac{1}{K \cdot r^{cdt}_i} \sum^{K}_{k=0} {\| (att_{i, k}^x, att_{i, k}^y), (\widetilde{c^x_i}, \widetilde{c^y_i})   \|_2}
	\end{equation}
	The probability points in one GT region need to be as few as possible. 
	To take the number of points into consideration, we define $R_{da}$ as:
	\begin{equation}
	R_{da} =
	\left\{
	\begin{array}{lr}
	0, & \ell = 0. \\
	1, & \ell = 1. \\
	\frac{1}{N} \sum^{N}_{i}   \frac{ (1 - d^{pd}_i) + exp( - \eta  (\ell + 2)) }{2}, & \ell \geq 2.\\
	\end{array}
	\right.
	\end{equation}
	where $\eta$ is the probability softening coefficient, which is leveraged to balance the influence of the probability number $\ell$ on the result.
	
	$R_{da}$ quantifies the number of probability points and the probability divergence distance in the GT region, thereby obtaining the degree of attention dispersion of the generated SeLo map.
	When there is only one probability point in the GT area, $R_{da}$ is at the maximum value of 1, while for other cases we quantify it to the interval 0-1.
	On the one hand, $R_{da}$ can be utilized to evaluate the stability of the retrieval model when performing SeLo task, while on the other hand, it can also judge whether the RS image contains any target or relationship in the query.

	\begin{figure*}[!t]
	\centering
	\includegraphics [width=7.2in]{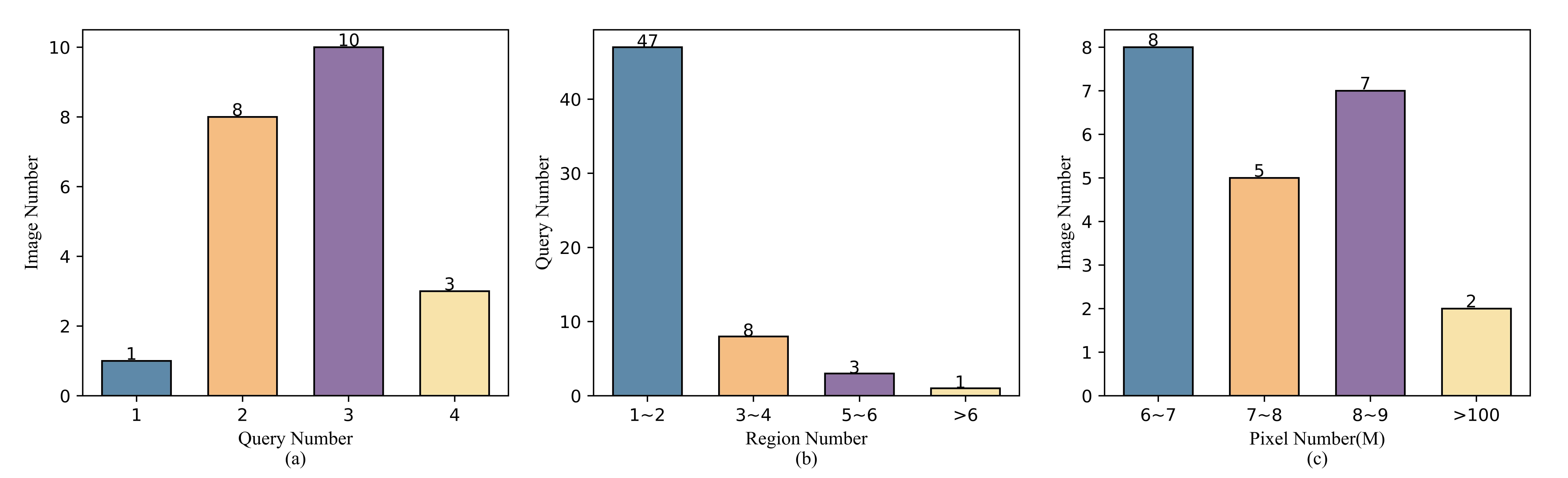}
	\caption{
		Data statistics of Semantic Localization Testset.
		(a) The correspondence between the query number and image number.
		(b) The correspondence between the region number and query number.    
		(c) The correspondence between the pixel number and image number.        
	}
	\label{data_info}
\end{figure*}

	\subsection{Establishment of unified indicators}
	The proposed indicators measure the SeLo task from different aspects, and only using one of them will not meet the scientific evaluation standard.
	After calculating the above metrics, the mean indicator $R_{mi}$ is generated to comprehensively measure the SeLo task:
	\begin{equation}
	R_{mi} = w_{su} R_{su} + w_{as}(1 - R_{as}) + w_{da}R_{da}
	\end{equation}
	where $w_*$ is the weighting factor, which is utilized to determine the importance of the proposed indicators.
	\textcolor{black}{$R_{su}$ reflects the attention ratio of the GT area and the non-GT regions, which is more important in the SeLo task, so we assign a larger weight to $R_{su}$.
	$R_{as}$, on the other hand, reflects the bias of attention. 
	Qualitatively speaking, this metric is more important than $R_{da}$, which measures the discreteness of attention.}
	Considering the importance of the indicators, we define $w_{su}$, $w_{as}$ and $w_{da}$ as 0.4, 0.35 and 0.25, respectively, thus giving the $R_{su}$ a larger proportion of the decision.
	\textcolor{black}{In addition, we also performed a manual comparison between this weight distribution method and the equal distribution method.
	By comparing multiple images, we found that this weight distribution method is more reasonable, which is also consistent with human perception.}
	$R_{mi}$ comprehensively measures various aspects of semantic localization, which provides a multivariate and reasonable quantitative score for the SeLo task.

	\section{Multi-modal Semantic Localization Testset}\label{sec-4}
	In this section, in order to evaluate SeLo task reasonably, we contribute a diverse semantic localization testset (AIR-SLT).
	We first introduce the annotation motivation for SeLo task, and then we summarize the data attributes of AIR-SLT.

	\subsection{How to Design Pertinent Semantic Localization Test Samples?}
	Our initial thought is to provide a comprehensive testset for SeLo task.
	However, what is a good SeLo test sample?
	To answer this question, we first take a step back to the motivation of SeLo task, which aims to locate the corresponding areas according to query text.
	If we are able to outline these areas manually, we will be able to preliminarily evaluate the generated SeLo maps.
	However, as SeLo task contain semantic-level information, we can not imitate existing tasks such as object detection and semantic segmentation to provide objective-level or pixel-level annotations.
	In contrast, we try to merely provide approximate labels which summarize the entities and relationships in the query for evaluation.
	Next, we evaluate the representation of models in color, relationship, and objects, meanwhile avoiding the semantic ambiguity as much as possible.

	During labeling, we try to include every query entity on the label as much as possible.
	In order to reflect the relationship between entities, we also add the region between entities to the ground truth.
	In regard to entities in the query that appeared in the images, we only annotate the parts that are semantically relevant to the query.
	The attention of SeLo may only focus on a point, thus providing relatively large bounding boxes seems to be meanless.
	However, this is the only way to quantify the attention areas, such annotations provide computational metrics that are positively correlated with actual accuracy.
	When the attention areas concentrate on GT areas, the calculated metrics will be higher, and vice versa, which indicates the reasonability of evaluation on SeLo task.

	\begin{table}[!t]
		\centering
		\caption{Quantitative Statistics of Semantic Localization Testset.}
		\begin{tabular}{cc|cc}
			\toprule
			Parameter & Value & Parameter & Value \\
			\midrule
			Word Number & 160   & Caption Ave Length & 11.20 \\
			Sample Number & 59  & Ave Resolution Ratio (m) & 0.3245 \\
			Channel Number & 3     & Ave Region Number & 1.75 \\
			Image Number & 22    & Ave Attention Ratio & 0.068 \\
			\bottomrule
		\end{tabular}%
		\label{tab:data_info}%
	\end{table}%

	\begin{table}[!t]
		\centering
		\caption{Different Class Number of Semantic Localization Testset.}
		\begin{tabular}{cc|cc|cc}
			\toprule
			Class & Number & Class & Number & Class & Number \\
			\midrule
			football field & 10    & lake  & 4     & trains & 1 \\
			parking lot & 8     & bridges & 2     & park  & 1 \\
			residential  & 7     & tenniscourts & 2     & forest & 1 \\
			plane & 7     & mountains & 1     & pond  & 1 \\
			baseball field  & 6     & bareland & 1     & beach & 1 \\
			ships & 5     & storagetank & 1     & -     & - \\
			\bottomrule
		\end{tabular}%
		\label{tab:cate}%
	\end{table}%

	\subsection{AIR-SLT: A New Testset for Semantic Localization}
	We contribute a semantic localization testset to provide systematic evaluation for SeLo task.
	The images in AIR-SLT come from Google Earth, and Fig. \ref{sample} exhibits several samples from the testset.
	Every sample includes a large RS image with the size of $3k\times2k$ to $10k\times10k$, a query sentence, and one or more corresponding semantic bounding boxes.
	During labeling, we add relationship information to the test cases, such as the orientation and distribution between two targets.
	The data information of AIR-SLT is shown in TABLE \ref{tab:data_info}.
	We provide 22 large-scale RS images with 3 channels, whose average resolution is 0.3245m.
	There are 59 test samples in total, in which the number of query words is 160, and the average length of the query is 11.20.
	The test cases consist of 17 different categories and the statistics of categories can be found in TABLE \ref{tab:cate}.
	The average number of GT boxes in every test case is 1.75, and the average proportion of attention ratio is 0.068.

	Besides, the statistics of the query in samples are shown in Fig. \ref{data_info}(a-b).
	For most samples, the number of queries is 2-3.
	Most queries only relate to 1-2 GT area, and there are a few queries consisting of multiple boxes due to semantic ambiguity.
	The pixel statistics (H$\times$W) of every image are shown in Fig. \ref{data_info}(c), ranging from 6M ($3k\times2k$) to 100M ($10k\times10k$), which suggests that the AIR-SLT has high-quality samples with huge size.
	AIR-SLT aims to provide a proper collection of test samples, thus we can evaluate the SeLo performance of retrieval models comprehensively.

	\begin{figure*}[!h]
	\centering
	\includegraphics [width=6.8in]{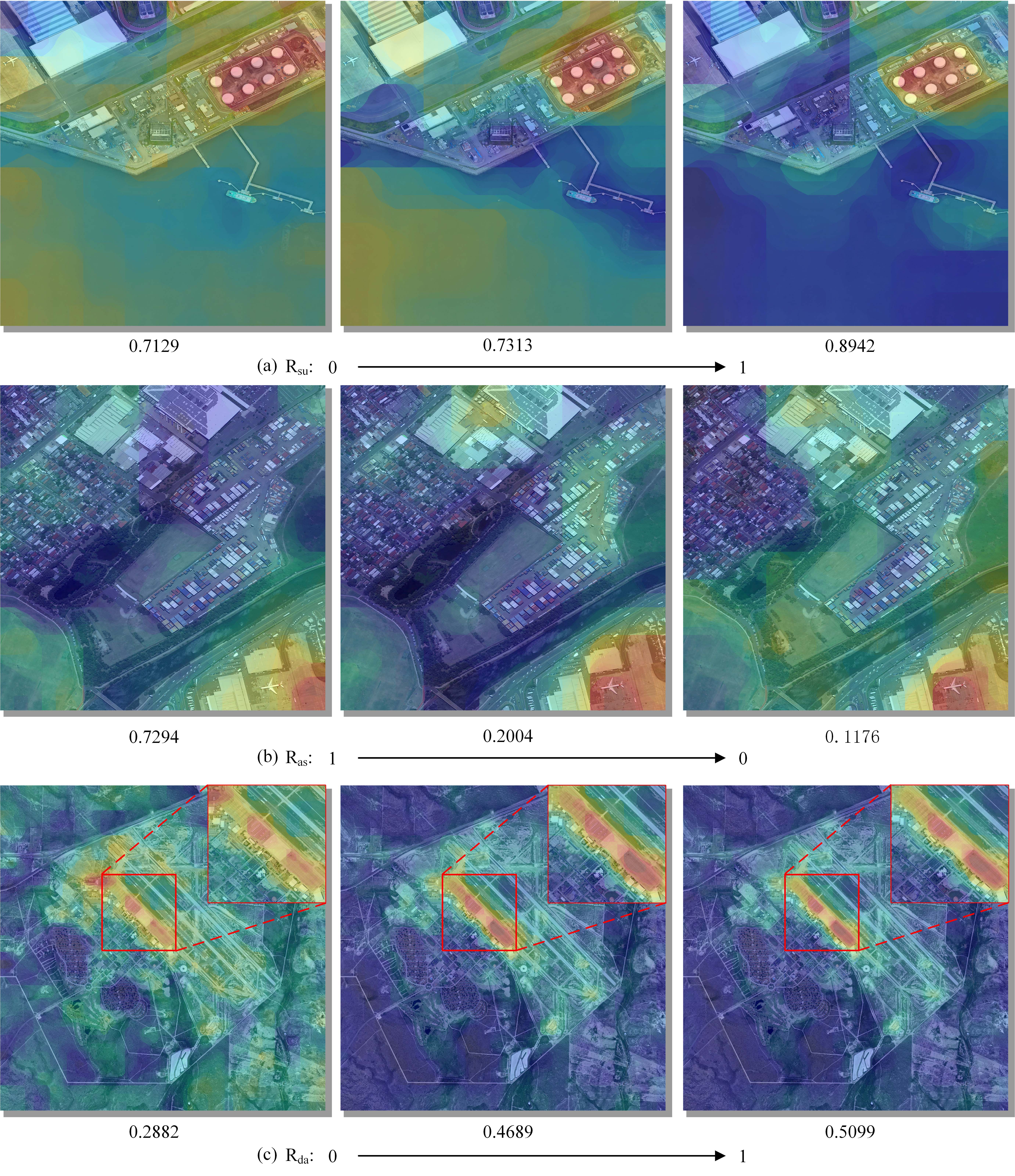}
	\caption{
		Qualitative analysis of SeLo indicator validity.
		(a) Query: ``eight large white oil storage tanks built on grey concrete floor''.
		(b) Query: ``a white plane parked in a tawny clearing inside the airport''.    
		(c) Query: ``lots of white and black planes parked inside the grey and white airport''.        
	}
	\label{indicator_verify}
\end{figure*}
	
	\section{Experiments Results and Analysis}\label{sec-5}
	
	In this section, we systematically conduct experiments on semantic localization, providing sufficient data support and theoretical analysis for this task.

	\subsection{Implementation Details}
	All experiments are performed on Intel(R) Xeon(R) Gold 6226R CPU @2.90GHz and a single NVIDIA RTX 3090 GPU.
	When we perform multi-scale cropping, except for declarations, we apply three scales of 256, 512, and 768 in default, and set the shift value as 0 and 0.5.
	\textcolor{black}{We set the $eps$ in $R_{su}$ as 1e-7 and set the parameter $\alpha$ as 0.694 in calculating the significant area proportion to make the critical threshold of $R_{su}$ reach under a uniform distribution.}
	The expansion factor $\Im$ of candidate regions is set to 1.5, allowing the model to take account of the nearby probability points comprehensively.
    \textcolor{black}{The scale $\beta$ in $R_{as}$ is set to 3 to achieve computational stability of the attention offset near the GT center.
    The probabilistic softening coefficient $\eta$ in $R_{da}$ is set to 0.5 to ensure the influence of the attention number to $R_{da}$.}

	\begin{figure*}[!h]
		\centering
		\includegraphics [width=6.8in]{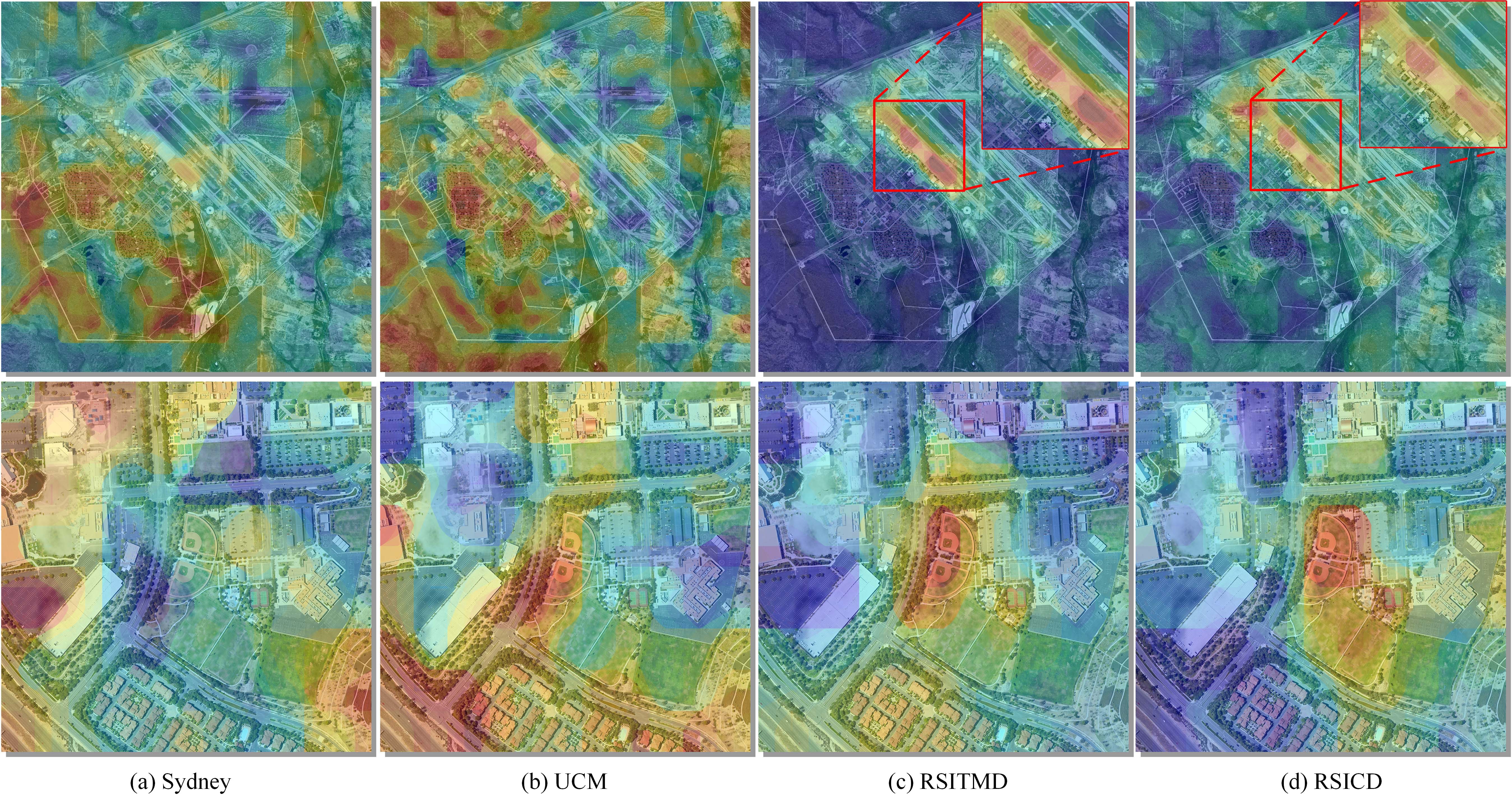}
		\caption{
			Qualitative comparison of SeLo performance on different datasets.
			(a) Query: ``lots of white and black planes parked inside the grey and white airport''.
			(b) Query: ``two green baseball fields next to two green soccer fields''.    
		}
		\label{diff_dataset}
	\end{figure*}

	\subsection{Indicator Effectiveness Analysis}
	In this subsection, we verify and analyze the effectiveness of the proposed multiple indicators.
	As shown in Fig. \ref{indicator_verify}, we provide three sets of examples to demonstrate the changes in SeLo performance along with the changes of $R_{su}, R_{as}$ and $R_{da}$.

	$R_{su}$ reflects the proportion of attention in GT areas.
	Fig. \ref{indicator_verify}(a) exhibits the $3k\times3k$ SeLo map using the query ``eight large white oil storage tanks built on grey concrete floor''.
	Along with the increase of $R_{su}$, the models put more attention on GT areas and decrease the attention proportion on other areas.
	\textcolor{black}{In Fig. \ref{indicator_verify}(c), although the attention in the GT region seems to decrease, the attention in the non-GT region greatly shrinks at this time.
	From the middle to the right of Fig.7(a), it can be found that the attention shrinkage in the non-GT region is extremely large.
	When the attention of the non-GT region greatly shrinks, even if the attention of the GT region remains unchanged, the quality of the generated SeLo map will still be improved relatively, as the most ideal situation is that the attention of the non-GT region is reduced to zero.
    In other words, if the attention of the GT and the non-GT regions is consistent and both at the maximum value, it is also impossible to obtain valid information from the SeLo map at this time.
    In summary, $R_{su}$ reflects the relative ratio of the both, rather than simply evaluating the attention on the GT region, thus making it a relatively important indicator in the SeLo task. }

	$R_{as}$ indicates the shift between the attention generated by the model and the center of the GT region.
	Fig. \ref{indicator_verify}(b) shows the retrieval results with the query ``a white plane parked in a tawny clearing inside the airport''.
	When $R_{as}$ is 0.7294, the attention of SeLo is obviously shifted, the reason is that the model asserts the regional features are matched at a large scale, but fails to perceive on small scale due to scale changes.
	After the model adjusts the attention, $R_{as}$ decreases to 0.2004 gradually.
	When we utilize multi-scale models to perform SeLo task, the attention of the models all focuses on GT areas, which obtains the lowest $R_{as}$ score with 0.1176.
	\textcolor{black}{$R_{as}$ reflects the degree of deviation between the attention and the target. 
	The closer the attention and the target are, the lower the value is.
	In this case, the attention center is still deviated from the GT center, so a small value exists.}

	$R_{da}$ reflects the discrete degree of attention in GT areas, which can be used to verify the stability of the model in retrieval.
	In Fig. \ref{indicator_verify}(c), we show the retrieval results from $10k\times10k$ RS scene with the query of ``lots of white and black planes parked inside the grey and white airport''.
	When the value of $R_{da}$ equals to 0.2882, although the models locate the planes in the airport, the attention is discrete, and there are missing targets.
	After optimization, false detections are largely eliminated, but the attention still drifted, which obtains a score of 0.4689 in $R_{da}$.
	After further removal of attention drift, the attention of the model successfully focused on the GT area, resulting in a $R_{da}$ score of 0.5009.
	\textcolor{black}{In this case, the model divides the attention into two parts due to the incoherence of the target.
	Therefore, at this time, A$R_{da}$ converges at about 0.5 instead of 1, which is still positively correlated with attention convergence degree.}

	Regarding the above three indicators, each of them performs a discriminative analysis on the generated SeLo results from different perspectives.
	From the above three cases, the proposed indicators can reasonably and perfectly quantify the generated SeLo map.
	The samples in Fig. \ref{indicator_verify}(c) also solidly reflect the effectiveness of the SeLo task for cross-modal semantic retrieval in large RS scenes.

	\subsection{Comparison of SeLo Performance of Different Trainsets}
	The data quality of the trainset has a significant impact on the SeLo performance.
	In \cite{A Lightweight Multi-Scale}, the authors qualitatively compared the SeLo performance on different RS image-text datasets and concluded that RSITMD is a great choice for this task.
	\textcolor{black}{In this subsection, we comprehensively compare the SeLo performance by training model on different datasets.}
	The datasets for comparison are as follows:
	
	\begin{itemize}
		
		\item \textbf{Sydney \cite{Deep semantic understanding}}: Sydney is the earliest image-text matching dataset, which contains only 555 samples, with an image and five sentences in each sample.
		\item \textbf{UCM \cite{Deep semantic understanding}}: UCM contains 2100 images of 21 scenes, with the size of 256$\times$256 pixels. The images in the UCM are from the National Map of the United States Geological Survey (USGS).
		\item \textbf{RSITMD \cite{Exploring a fine-grained}}: RSITMD provides 4k fine-grained image-text samples, this dataset has the lowest intra-class similarity and the highest number of words and categories in RS.
		\item 	\textbf{RSICD \cite{Exploring models and}}: RSICD includes 10k image-text pairs, which is currently the largest RS image-text dataset.
	\end{itemize}
	We use AMFMN to train on the above dataset and then evaluate the SeLo performance under different parameters.

	\begin{table}[!t]
		\centering
		\caption{Quantitative Comparison of SeLo Performance of Different Trainsets}
		\begin{tabular}{c|cccc}
			\toprule
			\multirow{2}[2]{*}{Dateset} & \multicolumn{4}{c}{Indicator} \\
			& $\uparrow$ $R_{su}$  & $\uparrow$ $R_{da}$  & $\downarrow$ $R_{as}$  & $\uparrow$ $R_{mi}$ \\
			\midrule
			Sydney & 0.5844  & 0.5670  & 0.5026  & 0.5496  \\
			UCM   & 0.5821  & 0.4715  & 0.5277  & 0.5160  \\
			RSITMD & \textbf{0.6920} & \textbf{0.6667} & \textbf{0.3323} & \textbf{0.6772} \\
			RSICD & 0.6661  & 0.5773  & 0.3875  & 0.6251  \\
			\bottomrule
		\end{tabular}%
		\label{tab:datasets}%
	\end{table}%

	The compare results are shown in Table \ref{tab:datasets}.
	Comparing the models trained on the Sydney and UCM datasets, it can be found that there is not much difference in $R_{su}$ indicator between them.
	The reason can be that these two datasets provide fewer training samples, and due to the small number of categories, the model lacks the understanding of words when performing the SeLo task.
	Compared with them, RSICD provides more samples, which greatly reduces the representation bias of the model for multi-modal samples.
	However, due to the strong intra-class similarity of RSICD, the model suffers from positive sample ambiguity during training, which reduces the performance of the model on the SeLo task.
	Compared with all the other datasets, RSITMD provides more fine-grained samples and also has a larger number of words, which can better satisfy the retrieval needs.
	Therefore, in terms of quantitative indicators, RSITMD is indeed a relatively complete RS image-text cross-modal dataset in SeLo task.

	As shown in Fig. \ref{diff_dataset}, we simultaneously conduct a qualitative comparison and analysis of retrieval models trained on different datasets.
	The query used in the upper sample of Fig. \ref{diff_dataset} is ''lots of white and black planes parked inside the grey and white airport'', and the size of the RS image for retrieval is $10k\times10k$ pixels.
	For the models trained on Sydney and UCM datasets, the attention is scattered when performing the SeLo task, and fails to focus on the region of interest.
	For RSICD, although the airport and the planes are located, the model gives false predictions on details of the upper left of the airport.
	Compared to the above three datasets, the model trained on RSITMD has the neatest attention, which verifies the high quality of RSITMD dataset.
	The size of the bottom sample in Fig. \ref{diff_dataset} is $2304\times2816$, and the query is ``Two green baseball fields next to two green soccer fields''.
	Compared to the results from small datasets like Sydney and UCM, the models trained on RSITMD and RSICD achieve better visual results.
	The attention of the model trained on RSICD is also scattered around the GT regions compared to RSITMD.
	The above experiments once again demonstrate the great advantages of RSITMD in image-text training and actual deployment.

	\begin{figure*}[!t]
		\centering
		\includegraphics [width=6.8in]{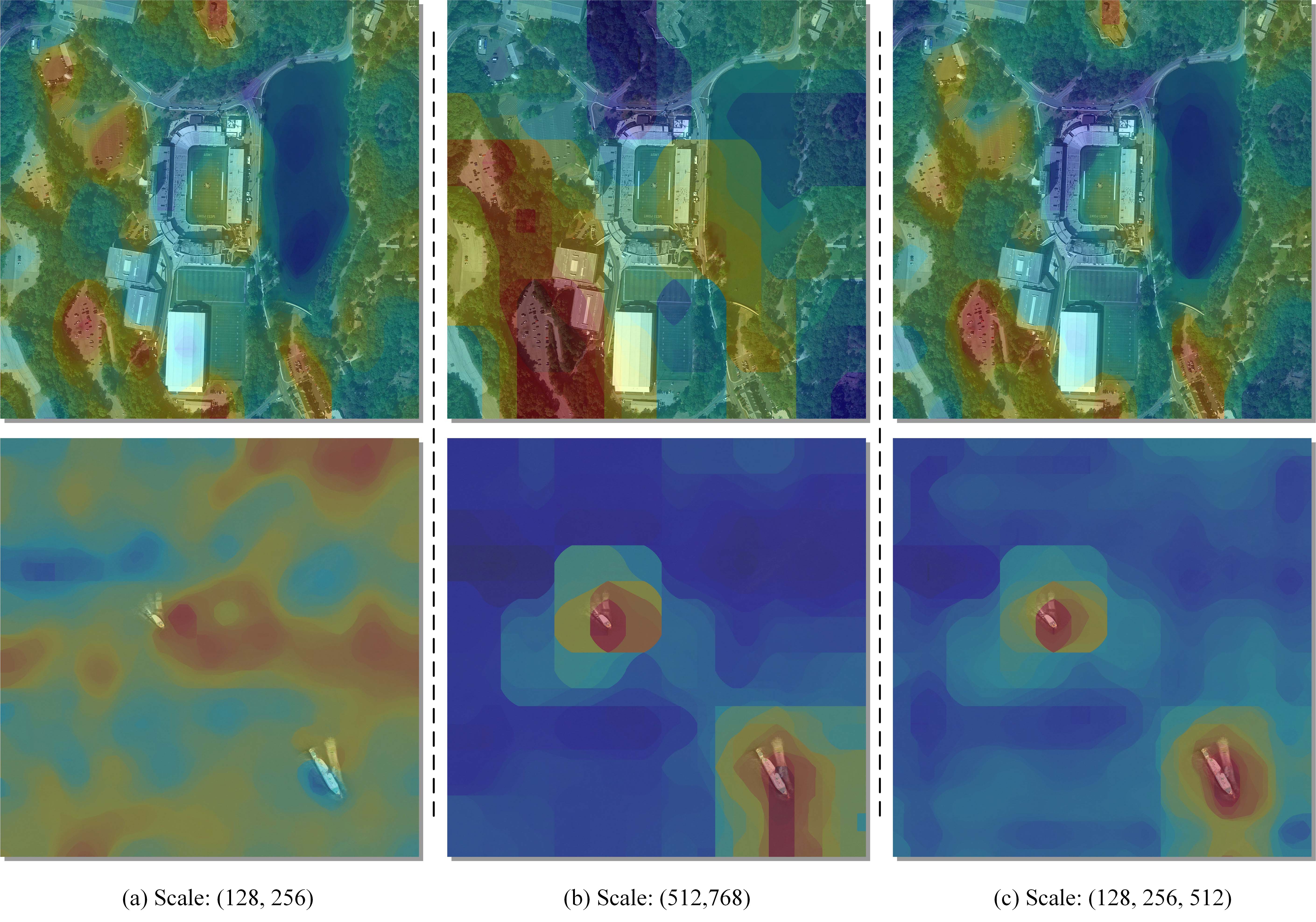}
		\caption{
			Qualitative comparison of SeLo performance on different scales.
			(a) Query: ``some cars are parked in a parking lot surrounded by green woods''.
			(b) Query: ``one white boat sailing on the dark blue sea''.    
		}
		\label{diff_scale}
	\end{figure*}

	\begin{table*}[!t]
		\centering
		\caption{Quantitative Comparison of SeLo Performance on Different Scales}
		\begin{tabular}{c|cccc|cccc|c}
			\toprule
			& \multicolumn{4}{c|}{Scale}    & \multicolumn{4}{c|}{Indicator} & \multirow{2}[2]{*}{Time(m)} \\
			& 128   & 256   & 512   & 768   & $\uparrow$ $R_{su}$  & $\uparrow$ $R_{da}$  & $\downarrow$ $R_{as}$  & $\uparrow$ $R_{mi}$  &  \\
			\midrule
			s1    & \checkmark     & \checkmark     &       &       & 0.6389  & 0.6488  & 0.2878  & 0.6670  & 33.81  \\
			s2    &       & \checkmark     & \checkmark     &       & 0.6839  & 0.6030  & 0.3326  & 0.6579  & 14.25  \\
			s3    &       &       & \checkmark     & \checkmark     & 0.6897  & 0.6371  & 0.3933  & 0.6475  & \textbf{11.23} \\
			s4    & \checkmark     & \checkmark     & \checkmark     &       & 0.6682  & \textbf{0.7072} & \textbf{0.2694} & \textbf{0.6998} & 34.60  \\
			s5    &       & \checkmark     & \checkmark     & \checkmark     & \textbf{0.6920} & 0.6667  & 0.3323  & 0.6772  & 16.92  \\
			s6    & \checkmark     & \checkmark     & \checkmark     & \checkmark     & 0.6809  & 0.6884  & 0.3025  & 0.6886  & 36.28  \\
			\bottomrule
		\end{tabular}%
		\label{tab:scale}%
	\end{table*}%

	\subsection{Comparison of SeLo Performance on Different Scales}
	In the SeLo task, the cropping scale is used as a preprocessing hyperparameter, which determines the size of the model's receptive field.
	When the cropping scale is too large, the model cannot predict more fine-grained results, and when the scale is too small, the model can fail to understand the global features of the image.
	In this subsection, we qualitatively and quantitatively analyze the performance and elapsed time of SeLo task at different cropping scales, where elapsed time refers to the time to generate all SeLo maps and the time to evaluate.
	
	As shown in Table \ref{tab:scale}, we conduct 6 sets of ablation experiments (s1 - s6) to explore the changes in performance and elapsed time at different cropping scales.
	For the experiment s1, since we only apply a small-scale sliding window, more slices requiring the similarity calculation are generated, which is much more time-consuming than the experiments s2 and s3.
	Compared with s1, the scales of s2 and s3 are more coarse-grained, which reduces the $R_{mi}$, but takes less time.
	As for s4, which adds a scale of 512 on the basis of s1,  obtains the highest $R_{mi}$.
	Compared with the above-mentioned experiments, s5 discards the 128 scales which occupied for a long time, and although the performance decreases, it obtains a great time gain.
	Interestingly, the performance of SeLo does not continue to grow compared to s4 when all scales are evaluated simultaneously, which suggests that large-scale slices are not as effective for this task.
	We use the s5 scale by default in practice, which not only obtains a faster generation speed, and also basically maintains the performance of SeLo.

	\begin{figure*}[!h]
		\centering
		\includegraphics [width=6.8in]{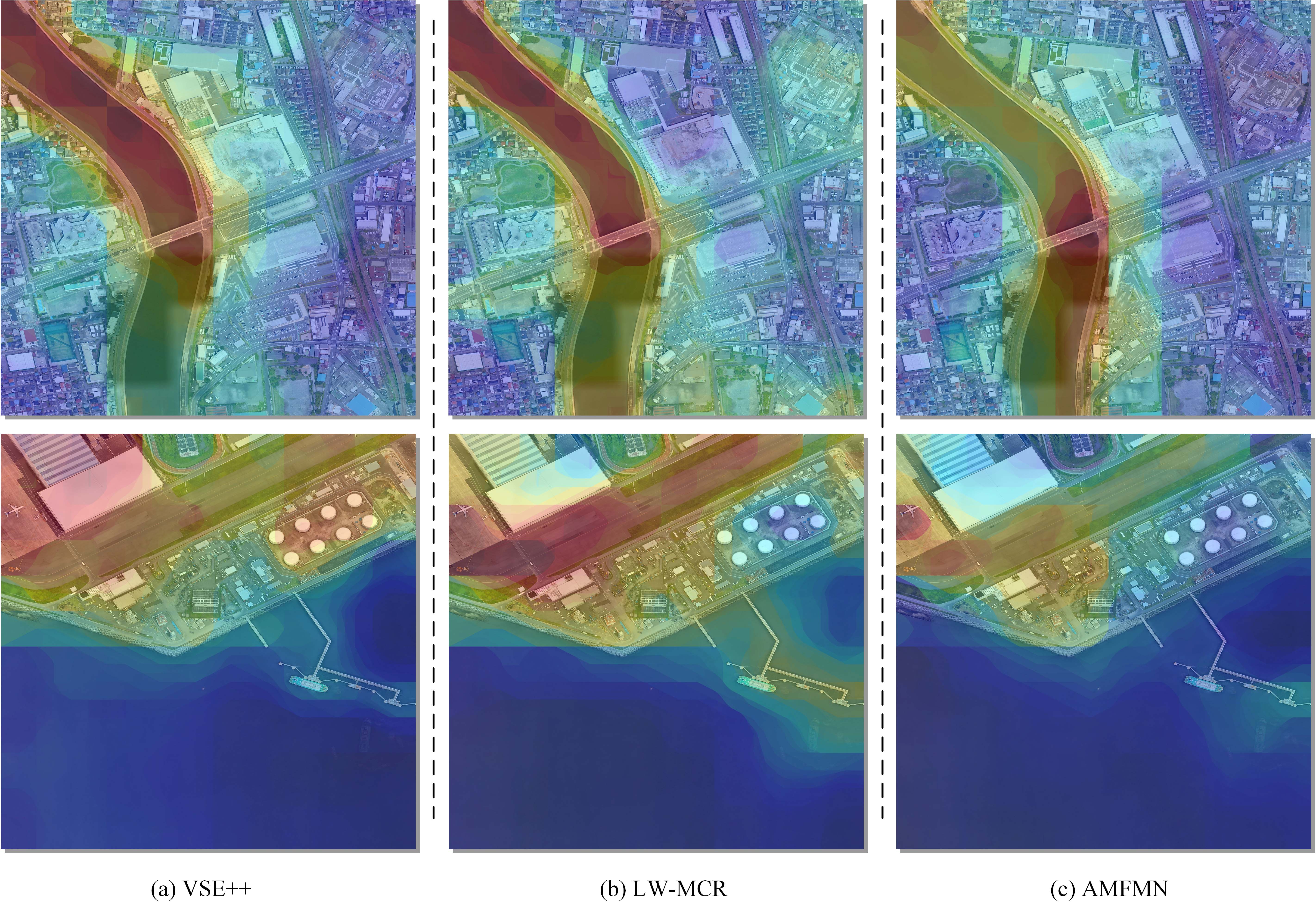}
		\caption{
			Qualitative comparison of SeLo performance on different models.
			(a) Query: ``a bridge spans a dark blue river''.
			(b) Query: ``a white plane parked on a gray airport''. 
		}
		\label{diff_model}
	\end{figure*}

	As shown in Fig. \ref{diff_scale}, we visualize the qualitative results of the generated SeLo maps with different scales.
	The size of the image in top of Fig. \ref{diff_scale} is $2304\times3072$ and the corresponding query is "some cars are parked in a parking lot surrounded by green woods".
	In such a test case, the parking lot is small, so utilizing a small scale produces a better SeLo map than using a large scale.
	The size of the image below Fig. \ref{diff_scale} is $3000\times3000$, and the query corresponding to this image is "one white boat sailing on the dark blue sea".
	In this example, when only small scales are used for retrieval, the model fails to recognize the relationship between entities in the query, resulting in strong false positives.
	For larger scales, the SeLo map generated at this time has a better effect.
	To sum up, there is no unified conclusion about the choice of size and scale, and specific analysis should be carried out according to specific tasks.
	However, the comprehensive scales often achieve better results, which is also an important reason why we choose the multiple scales as the default scale.

	\begin{table}[!t]
		\centering
		\caption{Quantitative Comparison of SeLo Performance on Different Retrieval Models}
		\begin{tabular}{c|cccc|c}
			\toprule
			\multirow{2}[1]{*}{Models} & \multicolumn{4}{c|}{Indicator} & \multirow{2}[1]{*}{Time(m)} \\
			& $\uparrow$ $R_{su}$  & $\uparrow$ $R_{da}$  & $\downarrow$ $R_{as}$  & $\uparrow$ $R_{mi}$  &  \\
			\midrule
			VSE++ & 0.6364  & 0.5829  & 0.4166  & 0.6045  & 15.61 \\
			LW-MCR & 0.6698  & 0.6021  & 0.4335  & 0.6167  & \textbf{15.47} \\
			SCAN  & 0.6421  & 0.6132  & 0.3871  & 0.6247  & 16.32 \\
			CAMP  & 0.6819  & 0.6314  & 0.3912  & 0.6437  & 18.24 \\
			AMFMN & \textbf{0.6920} & \textbf{0.6667} & \textbf{0.3323} & \textbf{0.6772} & 16.92 \\
			\bottomrule
		\end{tabular}%
		\label{tab:model}%
	\end{table}%

	\subsection{Comparison of SeLo Performance on Different Retrieval Models}
	
		\begin{figure*}[!h]
		\centering
		\includegraphics [width=6.8in]{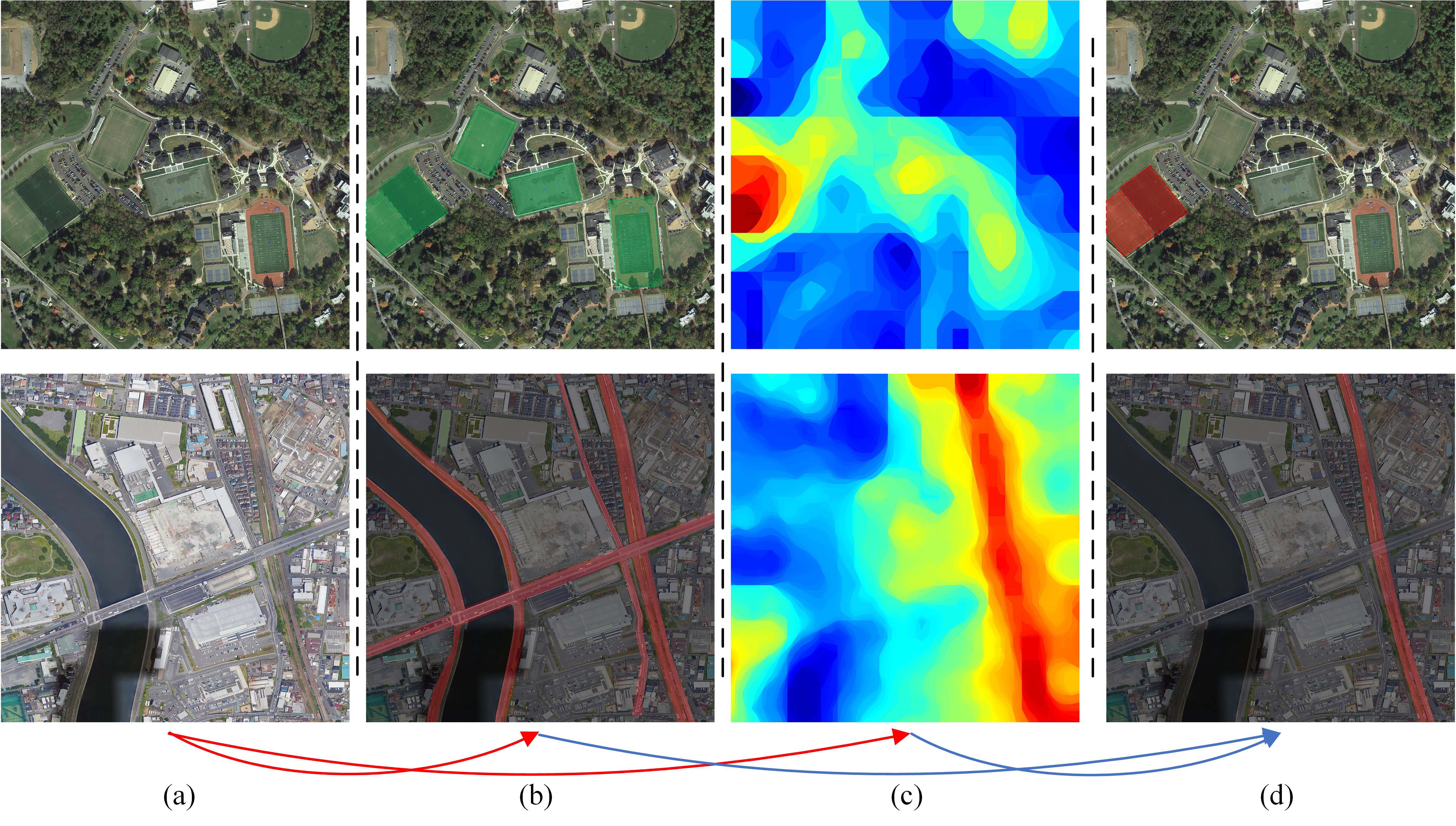}
		\caption{
			Combine SeLo with other tasks.
			The top of the figure shows the detection results after add the SeLo map with query of ``two parallel green playgrounds''.
			The bottom of the figure shows the road extraction results after add the SeLo map with query of ``the red rails where the grey train is located run through the residential area''.
			(a) Source images.
			(b) Results of specific tasks.
			(c) Results of specific SeLo maps.
			(d) Fusion results of specific tasks and SeLo map.
		}
		\label{selo_with_subtask}
	\end{figure*}

	In this section, \textcolor{black}{we provide various indicators of different RSCTIR models in SeLo task}, so as to bring sufficient data accumulation to other scholars.
	In the experiment, we apply the RSITMD dataset uniformly, and use the scales (256, 512, 768) for multi-scale sliding window.
	The model for this evaluation is as follows:
	\begin{itemize}
		\item \textbf{VSE++ \cite{VSE}}: VSE++ use convolution networks and recurrent networks to embed image and text information into the same space and proposes the triplet loss to \textcolor{black}{train the image-text retrieval model.}
		\item \textbf{LW-MCR \cite{A Lightweight Multi-Scale}}: LW-MCR achieves fast similarity calculation by discarding the cumbersome network framework through lightweight network modeling.
		\item \textbf{SCAN \cite{SCAN}}: The SCAN model, which is based on VSE++, uses region information extracted from image features and aligns the target in the image with the target in the text.
		\item \textbf{CAMP \cite{CAMP}}: The CAMP model proposes a method of adaptive message passing, which adaptively controls the flow of cross-modal information transmission and calculates the matching scores of images and texts by using fusion features.
		\item \textbf{AMFMN \cite{Exploring models and}}: AMFMN joins the MVSA mechanism to filter the useless information in the image to mask the proposed global visual information.	
	\end{itemize}
	We follow AMFMN to uniformly set the visual backone to ResNet-18 \cite{resnet}, and for SCAN we regard the features proposed by backbone as region features.
	
	As shown in TABLE \ref{tab:model}, we separately show the performance of these models on the SeLo task and list the time required for inference.
	Compared to VSE++, LW-MCR is less time-consuming, which verifies that the model has less computational complexity.
	Compared with VSE++, SCAN adds region alignment operations, so there is a corresponding increase in inference time.
	CAMP uses the method of modal fusion, the effect is relatively good, but not dominant in time.
	AMFMN reaches a high level with all indicators at the forefront of the SeLo task, \textcolor{black}{and the comprehensive indicator reaches 0.6772.}
	
	As shown in Fig. \ref{diff_model}, we show examples of SeLo maps generated by different models which trained on the RSITMD dataset.
	Image size of the top of Fig. \ref{diff_model} is $3000\times3000$, with query of ``a bridge spans a dark blue river''.
	When using VSE++ and LW-MCR for the SeLo task, the model pays attention to the river in addition to the bridge, which reflects that the model does not understand the relational properties of the query at this time.
	When utilizing AMFMN for the SeLo task, the model focuses most of the attention on the bridge, and the attention in the residential area is lower than that of the LW-MCR.
	The size of the image below Fig. \ref{diff_model} is also $3000\times3000$, and the query is ``a white plane parked on a gray airport''.
	For the models generated by the first two, in addition to the aircraft, the attention is also dispersed on the airport and the runway.
	The SeLo map generated by AMFMN locates the aircraft more accurately, and although the attention is a little scattered at the airport, it is still the best.
	In summary, the AMFMN with the MVSA mechanism performed well in the SeLo task.
	However, the current model still has a lot of room for improvement on the SeLo task, which requires researchers to invest more energy to promote this task.

	\subsection{Analysis of Time Consumption}

	In this subsection, we analyze the time occupancy in the SeLo task, which is divided into four phases:
	\begin{itemize}
		\item \textbf{Cut}: Segmentation by multi-scale sliding window.
		\item \textbf{Sim}: Slice similarity calculation.
		\item \textbf{Gnt}: Pixel-level similarity stacking.
		\item \textbf{Flt}: Noise filtering.
	\end{itemize}
	\textcolor{black}{We count the time occupancy ratios of the above four stages at different scales to analyze the time consumption in the SeLo task.
	When calculating the total time, we calculate the total time for generating and evaluating the SeLo map, so the total time proportion of the four stages is not 1.}
	Additionally, we set the batchsize to 1 in our tests for a fair comparison.

	\begin{table}[!t]
		\centering
		\caption{Quantitative Analysis of Time Consumption}
		\begin{tabular}{cccccc}
			\toprule
			\multicolumn{6}{c}{Scale (128, 256)} \\
			& Cut   & Sim   & Gnt   & Flt & Total \\
			Times(m) & 2.85  & 20.60  & 7.40  & 0.73  & 33.81  \\
			Rate(\%) & 8.42  & 60.94  & 21.88  & 2.16  & - \\
			\midrule
			\multicolumn{6}{c}{Scale (512, 768)} \\
			& Cut   & Sim   & Gnt   & Flt & Total \\
			Times(m) & 0.46  & 1.17  & 6.96  & 0.67  & 11.23  \\
			Rate(\%) & 4.06  & 10.42  & 61.98  & 5.97  & - \\
			\midrule
			\multicolumn{6}{c}{Scale (256, 512, 768)} \\
			& Cut   & Sim   & Gnt   & Flt & Total \\
			Times(m) & 0.93  & 5.72  & 7.38  & 0.74  & 16.92  \\
			Rate(\%) & 5.52  & 33.82  & 43.60  & 4.37  & - \\
			\bottomrule
		\end{tabular}%
		\label{tab:times}%
	\end{table}%

	The contrast results are as shown in TABLE \ref{tab:times}.
	When the scale is (128, 256), it can be seen that the slice time and similarity calculation time are higher due to more small slices.
	In this case, most of the time consumed by the model is spent on similarity computation.
	Pixel-level similarity stacking is also a time-consuming process, which requires pixel-level averaging of the similarities of all sub-slices.
	When using the scale (512, 768) for SeLo, since the number of generated slices is significantly reduced, the required cropping time and similarity calculation time are also significantly reduced.
	In this case, the similarity stacking method occupies most of the time loss, and at the same time, the time proportion of noise filtering is also greatly increased.
	When using scale (256, 512, 768) for multi-scale sliding windows, the time loss still focus on similarity calculation and probability map generation.
	It can be seen that in order to further reduce the time consumption, the Sim step and the Gnt step need to be further optimized.
	For the Sim step, time can be reduced by designing lightweight models and parallel computing.
	For the Gnt step, a more mature and efficient similarity mixing algorithm needs to be designed to enable fast probability map generation.
	We will further explore both in future work.

	\subsection{Exploring combine SeLo with Other Tasks: A New Paradigm for RS Referring Expression}
	
	Referring expression comprehension (REC) refers to localizing certain bounding boxes according to semantic text on a basis of detection \cite{MAttNet}\cite{A Real-Time Global}.
	This task requires dual supervised information for detection and text in natural scenes, which is difficult to obtain in the RS field.
	Furthermore, due to the small size of objectives in RS and strong intra-class similarity, it is difficult to carry out REC in this field.
	Considering the semantic properties of SeLo, we design a paradigm of SeLo joint subtasks to conduct REC tasks in the RS area.
													
	
	Fig. \ref{selo_with_subtask} shows the bounding boxes and segmentation regions when SeLo is combined with detection and road extraction from top to bottom, respectively.
	The top of Fig. \ref{selo_with_subtask} shows the detection results after adding the SeLo map with query of ``two parallel green playgrounds''.
	When only the detector is used to obtain the playground in the RS image, although good detection results can be obtained, when faced with semantic information such as two parallel playgrounds, the traditional method is powerless.
	However, for specific semantic information, SeLo can obtain the general location of semantic relations.
	Taking advantage of the respective advantages of the above tasks, the semantic information required for detection can just be compensated by the SeLo map, thus to get the detection bounding box with semantics.
	For the road extraction task, when specific road needs to be extracted, the SeLo map with semantics can be utilized for information complementation.
	The advantage of SeLo map lies in high-level semantics, which is not available for other tasks.
	However, due to the scarcity of paired data and the scale of sliding windows, SeLo currently cannot achieve pixel-level segmentation and detection tasks.
	We expect more scholars can focus on this field, by providing a large number of image-text samples or providing a strong RSCTIR pretrained model to greatly improve the small-scale resolution capability of SeLo.

	\section{Conclusion}\label{sec-6}
	
	This article systematically studies and explores the semantic localization task, which is a higher-level retrieval task compared to other tasks such as object detection and road extraction.
	To provide quantitative analysis of SeLo, we propose three reasonable indicators to systematically evaluate this task from different aspects.
	Furthermore, a semantic localization testset and extensive SeLo analysis are contributed to advance the SeLo task.
	Qualitative and quantitative experiments demonstrate that the SeLo task has great application value, which may be the next research hospot in RS retrieval tasks.
	
	\ifCLASSOPTIONcaptionsoff
	\newpage
	\fi

	\begin{IEEEbiography}[{\includegraphics[width=1in,height=1.25in,clip,keepaspectratio]{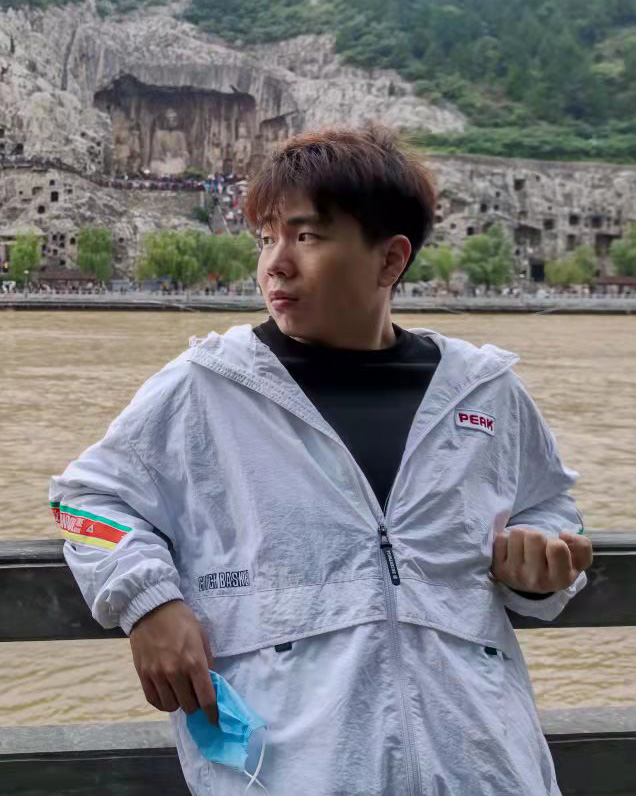}}]{Zhiqiang Yuan}
		received the B.Sc. degree from the Harbin Engineering University, Harbin, China, in 2019. He is currently pursuing the Ph.D. degree with the Aerospace Information Research Institute, Chinese Academy of Sciences, Beijing, China.
		
    	His research interests include computer vision, pattern recognition, and generative model, especially on remote sensing image processing.  (yuanzhiqiang19@mails.ucas.ac.cn)
	\end{IEEEbiography}

	\begin{IEEEbiography}[{\includegraphics[width=1in,height=1.25in,clip,keepaspectratio]{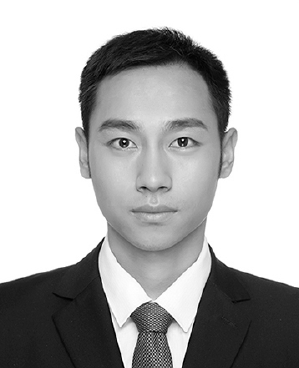}}]{Wenkai Zhang}
	recevied the B.Sc. degree from China University Of Petroleum, TsingTao, China, in 2013 and the Ph.D. from Institute of Electronics, Chinese Academy of ciences in 2018.
	
	He is currently an assistant professor in the Aerospace Information Research Institute, Chinese Academy of Sciences, China. His research interests include multi-modal signal processing, image segmentation, and pattern recognition. 
    \end{IEEEbiography}

	\begin{IEEEbiography}[{\includegraphics[width=1in,height=1.25in,clip,keepaspectratio]{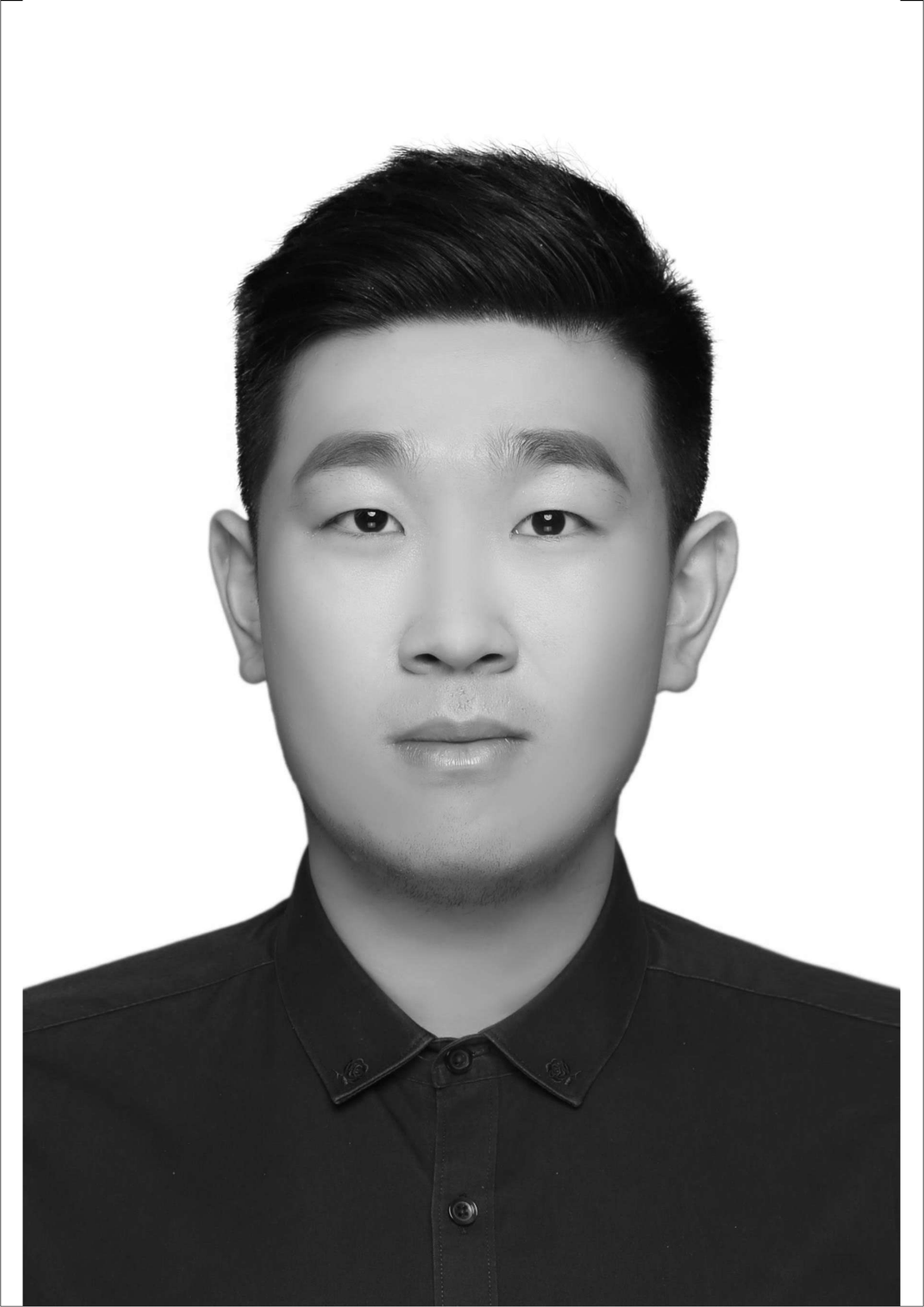}}]{Chongyang Li}
	 received the B.Sc. degree from the Dalian University of Technology, Dalian, China, in 2022. He is currently pursuing the Ph.D. degree with the  Aerospace Information Research Institute, Chinese Academy of Sciences, Beijing, China.
	
	 His research interests include multi-modal remote sensing image interpretation and multi-modal signal processing.
    \end{IEEEbiography}

	\begin{IEEEbiography}[{\includegraphics[width=1in,height=1.25in,clip,keepaspectratio]{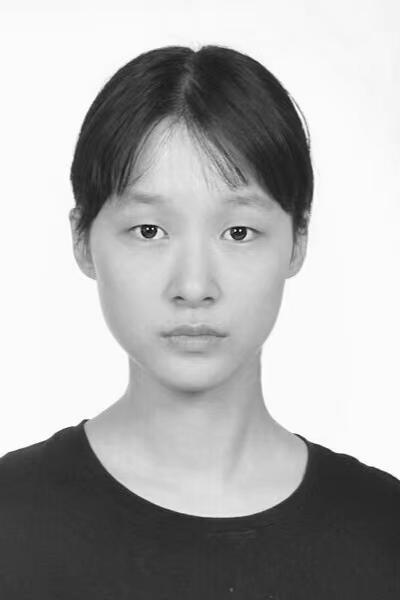}}]{Zhaoying Pan}
		received the B.Sc. degree from the University of Chinese Academy of Sciences, Beijing, China, in 2021. She is currently pursuing the M.S. degree in University of Michigan, Ann Arbor, United States.
		
		Her research interest includes multi-modal signal processing. 
	\end{IEEEbiography}

	\begin{IEEEbiography}[{\includegraphics[width=1in,height=1.25in,clip,keepaspectratio]{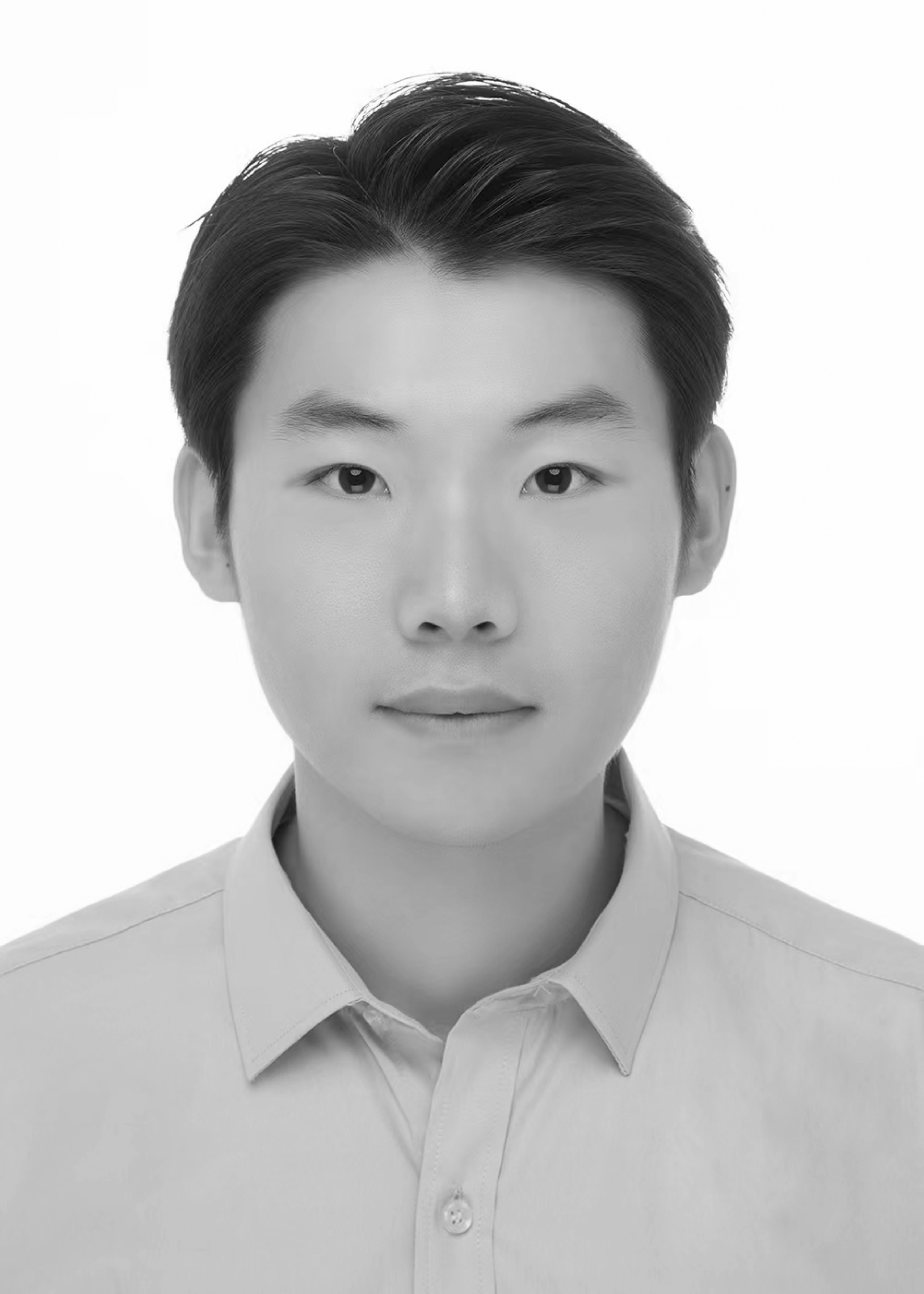}}]{Yongqiang Mao}
	Yongqiang Mao received the B.Sc. degree from Wuhan University, Wuhan, China, in 2019. He is pursuing the Ph.D. degree with the University of Chinese Academy of Sciences, Beijing, China, and the Aerospace Information Research Institute, Chinese Academy of Sciences, Beijing.

	His research interests include computer vision, pattern recognition, especially on 3D computer vision, semantic segmentation, etc.
	\end{IEEEbiography}

	\begin{IEEEbiography}[{\includegraphics[width=1in,height=1.25in,clip,keepaspectratio]{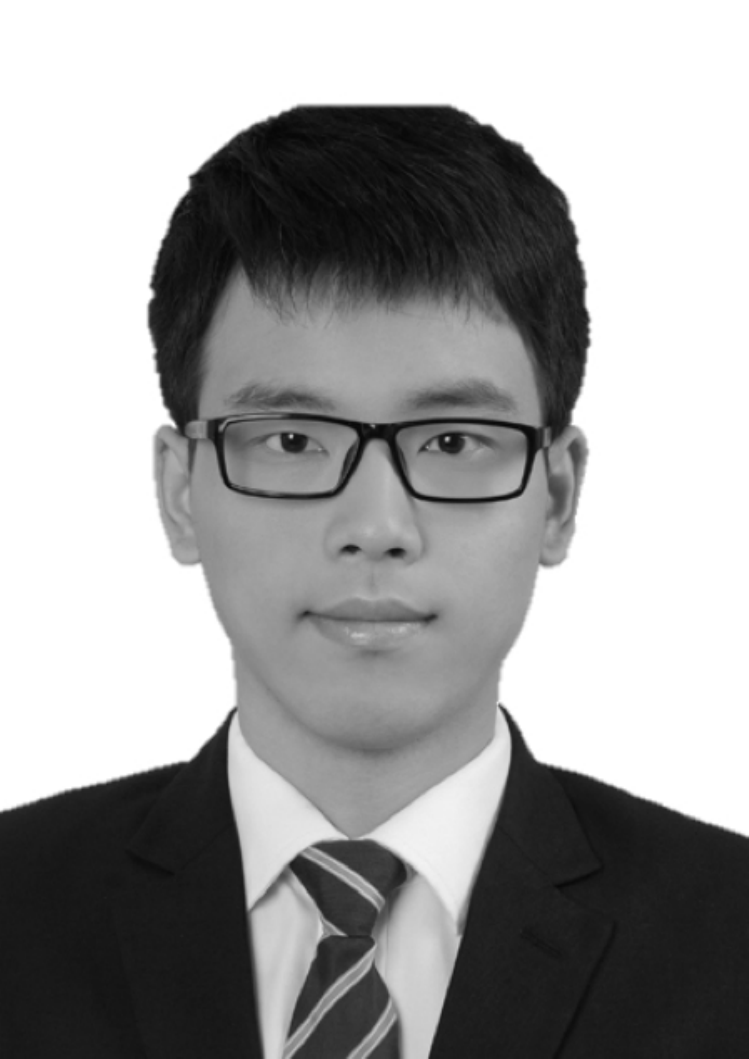}}]{Jialiang Chen}
		received the B.Sc. degree from the Zhengzhou University, Zhengzhou, China in 2012 and the M.Sc. degree from the Beijing Institute of Technology, Beijing, China in 2016.
	
		He is currently an assistant professor in the Aerospace Information Research Institute, Chinese Academy of Sciences, Beijing, China. His research interests include pattern recognition and remote sensing image processing.
	\end{IEEEbiography}

	\begin{IEEEbiography}[{\includegraphics[width=1in,height=1.25in,clip,keepaspectratio]{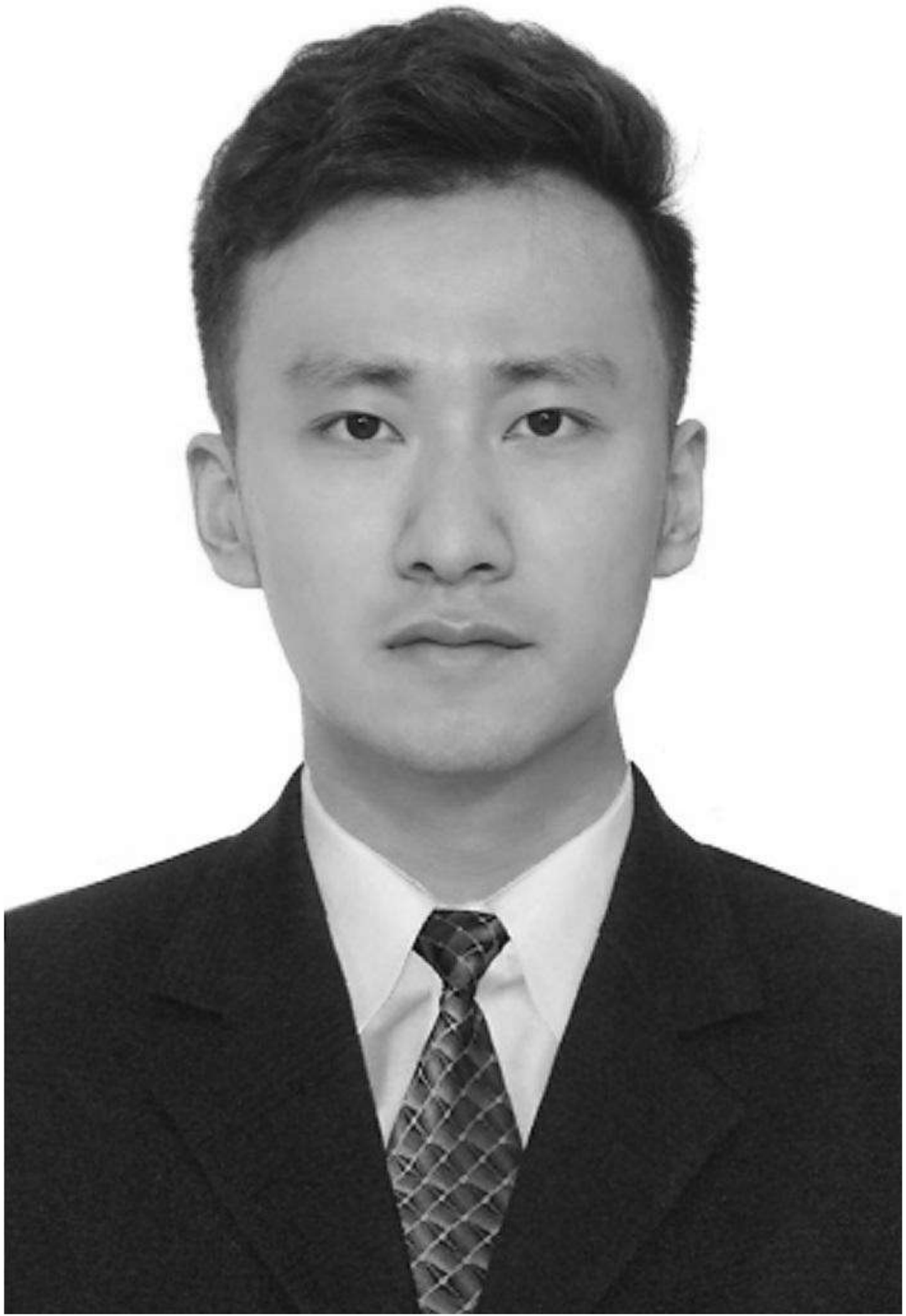}}]{Shuoke Li}
		recevied the B.Sc. degree from Xi'an Jiaotong University, Xi'an, China, in 2017 and the M.Sc. from Peking University, Beijing, China, in 2020. He is currently an assistant engineer in the Aerospace Information Research Institute, Chinese Academy of Sciences, China. 
		
		His research interests include multi-modal signal processing, image segmentation, and pattern recognition. 
	\end{IEEEbiography}

	\begin{IEEEbiography}[{\includegraphics[width=1in,height=1.25in,clip,keepaspectratio]{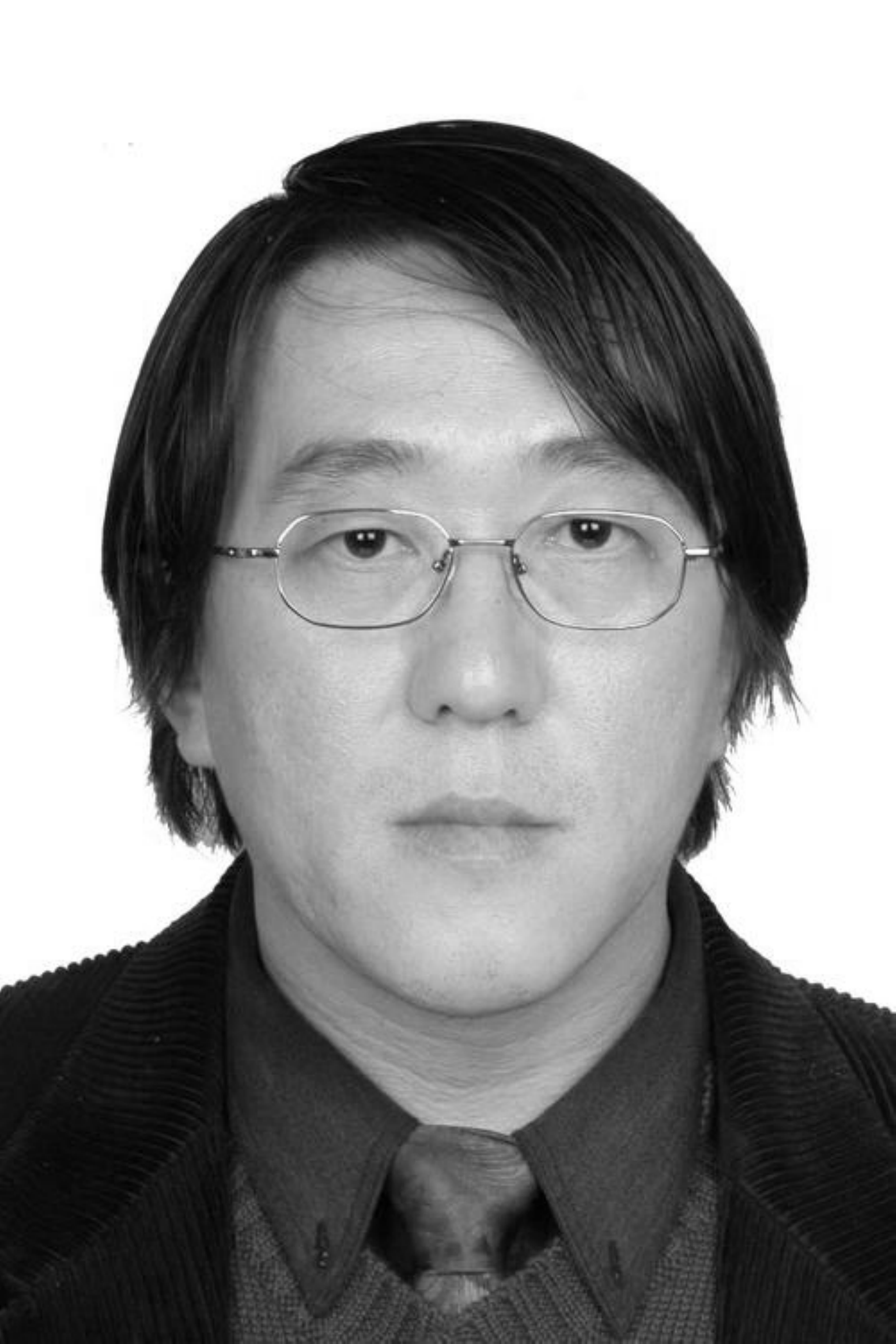}}]{Hongqi Wang}
		received the B.Sc. degree from the Changchun University of Science and Technology, Changchun, China, in 1983, the M.Sc. degrees from the Changchun Institute of Optics, Fine Mechanics and Physics, Chinese Academy of Sciences, Changchun, China, in 1988, and the Ph.D. degree from the Institute of Electronics, Chinese Academy of Sciences, Beijing, China, in 1994.
		
		He is currently a Professor with the Aerospace Information Research Institute, Chinese Academy of Sciences, China. 
	\end{IEEEbiography}
	
	\begin{IEEEbiography}[{\includegraphics[width=1in,height=1.25in,clip,keepaspectratio]{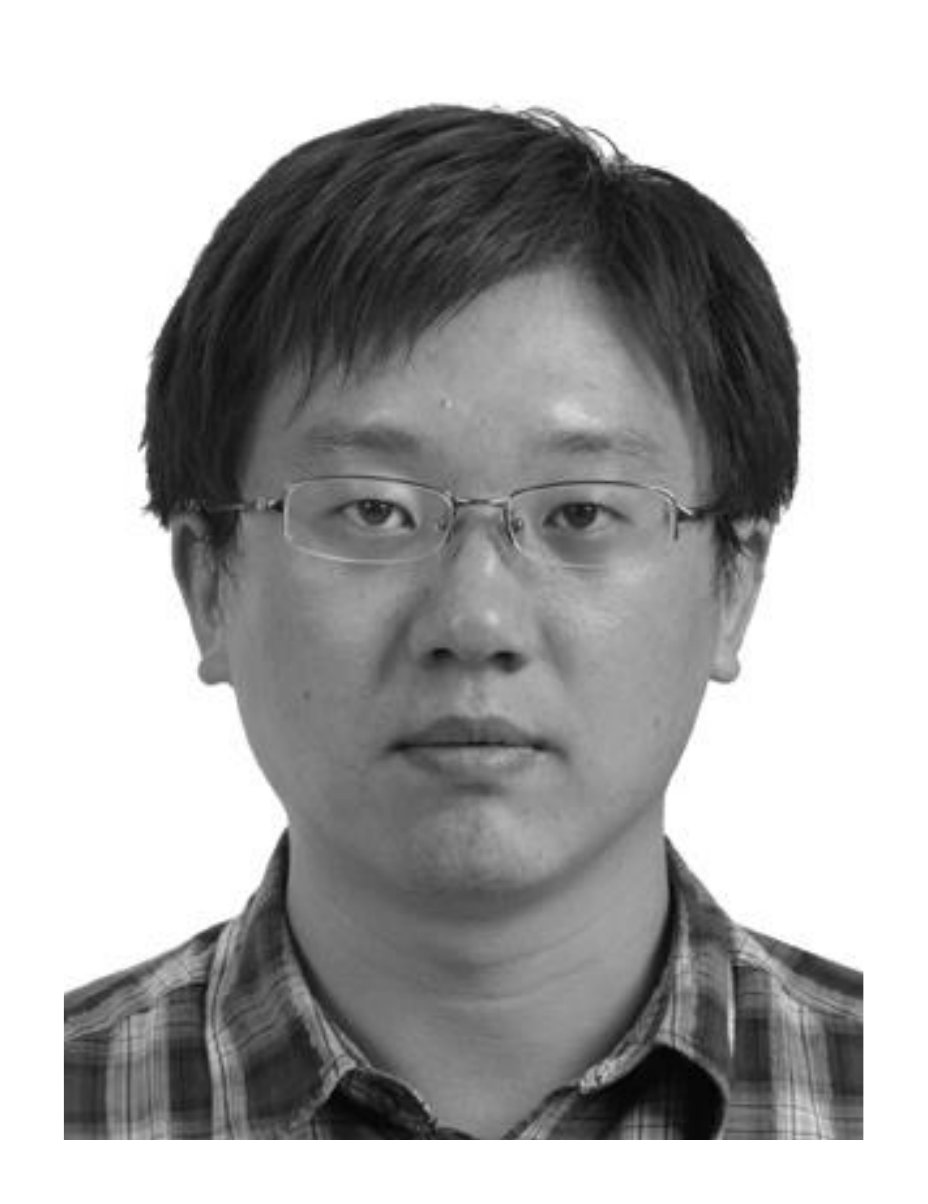}}]{Xian Sun}
		received the B.Sc. degree from the Beijing University of Aeronautics and Astronautics, Beijing, China, in 2004, and the M.Sc. and Ph.D. degrees from the Institute of Electronics, Chinese Academy of Sciences, China, in 2009.
		
		He is currently a Professor with the Aerospace Information Research Institute, Chinese Academy of Sciences, China. His research interests include computer vision, geospatial data mining, and remote sensing image understanding.
	\end{IEEEbiography}

\newpage

\begin{appendices}

\section{\textcolor{black}{\textbf{Questions \& Answers about \\ Semantic Localization}}}

\textcolor{black}{\textbf{Q: Why is the SeLo task based on retrieval, and is there a more lightweight and efficient way to do this task?}}

\textcolor{black}{A: We initially thought about multiple ways to perform SeLo tasks, including the following three methods:}
\begin{itemize}
	
	\item \textcolor{black}{The method based on referring expression comprehension (REC) [1][2], by establishing the relationship among the bounding boxes in RS, enables the model to obtain the semantic perception ability of the scene. 
	REC task usually uses a detector as a feed-forward network, then performs semantic-level modeling analysis based on the bounding boxes. 
	The REC task in natural scenes has been well-developed in the past few years. 
	However, in RS scenes, due to the complexity of small targets, this task cannot be carried out very well. 
	For example, when locating the relationship between the car and the house, the feedforward network first needs to obtain a large number of bounding boxes, and it is also a big challenge to obtain the perception of the house. 
	In addition, there are complete REC datasets in natural scenes, while REC annotations in RS scenes require a lot of further work to obtain. 
	The above conditions discouraged us from this solution, but we will continue to consider the feasibility of this method in future work.}
	
	\item \textcolor{black}{The method based on the adversarial generative approach (or can be understood as a method based on Class Activation Mapping (CAM) [3][4]), first came to our minds in late 2020. 
	Our initial idea was to utilize an end-to-end image generator, such as Unet [5], etc., to obtain the generated SeLo map under the condition of semantic embedding. 
	To supervise the above process, we use the framework for image-text similarity as the discriminator and maximize the similarity value between the masked RS image and the text, thus obtaining a well SeLo map. 
	Such a scheme seems to be feasible in natural scenes, and we have also witnessed the success of similar work in natural scenes during this time [6]. 
	However, when we implemented it at that time, we found that such a scheme was not ideal for small objects in large-scale scenes. 
	For images with tens of thousands of pixels, the direct end-to-end calculation is quite resource-intensive, and the performance of locating small objects is quite poor. 
	We will carry out further research on this scheme in the follow-up, hoping to make this scheme feasible with the development in the field.}
	
	\item \textcolor{black}{The method based on the similarity between images and texts, is also the scheme we are currently applying. 
	Despite the brute force essence, it is effective and feasible currently. 
	We decompose the SeLo task into sub-tasks of image-text similarity calculation and obtain the pixel-level cross-modal similarity through averaging to generate the SeLo map. 
	Such a scheme not only avoids the reliance on large datasets of referring expression comprehension, and also unbinds the model from the image size, which enables the model to generalize on data of different sizes. 
	Such a scheme is also due to the small scale of RS objects, which naturally has the advantage for the retrieval with multi-scale clipping, which is often not available in natural scenes with large objects.}
\end{itemize}

\textcolor{black}{Therefore, the SeLo framework can be regarded as a retrieval framework for small objects in large-scale scenes, which is extremely suitable to RS images.
The most ideal solution at present is to calculate the similarity with multi-scale clipping, however, there may be a better solution in the way of clipping.
For example, a front-end network can be designed to further reduce the amount of calculation through coarse-grained and fine-grained inspection.
Or filtering out statistically irrelevant scene slices through scene statistics may be another solution.
In fact,  there are also a lot to be improved, including the overlay of slicings, the processing of the overlap areas, and the post-processing of the SeLo map.
Despite these potential improvements, the aim of our paper is the explicit establishment of the entire framework, the proposal of the testset, and the establishment of the evaluation indicators.
We will continue to improve the retrieval framework on the basis of this paper in the follow-up work, hoping that we can provide some inspiration and accumulation for reseachers in related fields.}

\textcolor{white}{ }

\textbf{\textcolor{black}{References:}}

\textcolor{black}{[1] L. Yu et al., ``MAttNet: Modular Attention Network for Referring Expression Comprehension,'' 2018 IEEE/CVF Conference on Computer Vision and Pattern Recognition, 2018, pp. 1307-1315, doi: 10.1109/CVPR.2018.00142.}

\textcolor{black}{[2] M. Sun, W. Suo, P. Wang, Y. Zhang and Q. Wu, ``A proposal-free one-stage framework for referring expression comprehension and generation via dense cross-attention,'' in IEEE Transactions on Multimedia, doi: 10.1109/TMM.2022.3147385.}

\textcolor{black}{[3] P. -T. Jiang, C. -B. Zhang, Q. Hou, M. -M. Cheng and Y. Wei, ``LayerCAM: Exploring Hierarchical Class Activation Maps for Localization,'' in IEEE Transactions on Image Processing, vol. 30, pp. 5875-5888, 2021, doi: 10.1109/TIP.2021.3089943.}

\textcolor{black}{[4] T. Tagaris, M. Sdraka and A. Stafylopatis, "High-Resolution Class Activation Mapping," 2019 IEEE International Conference on Image Processing (ICIP), 2019, pp. 4514-4518, doi: 10.1109/ICIP.2019.8803474.}

\textcolor{black}{[5] X. He, Y. Zhou, J. Zhao, D. Zhang, R. Yao and Y. Xue, ``Swin Transformer Embedding UNet for Remote Sensing Image Semantic Segmentation,'' in IEEE Transactions on Geoscience and Remote Sensing, vol. 60, pp. 1-15, 2022, Art no. 4408715, doi: 10.1109/TGRS.2022.3144165.}

\textcolor{black}{[6] Xie, J., Hou, X., Ye, K., \& Shen, L. (2022). Cross Language Image Matching for Weakly Supervised Semantic Segmentation. arXiv preprint arXiv:2203.02668.}

\textcolor{white}{ }

\textbf{\textcolor{black}{Q: How to understand the scale issue in the SeLo task?}}

\textcolor{black}{
Scale is a rather important hyperparameter for the SeLo task.
Due to the large scale difference in RS scenes, using a single scale often cannot obtain well SeLo map.
However, such problems are not limited to SeLo tasks. 
For tasks such as detection and segmentation in large image, the single scale problem is also the difficulties that need to be overcome.
Furthermore, since the SeLo task extends individual similarity at the patch level, the window sliding method using multiple scales is essential to maintain robust pixel-level similarity.
}

\textcolor{black}{
The next thing to do is how to crop the image considering the size of different windows.
Although RS scenes have large scale differences, the RS images captured by most of the existing earth observation satellites cannot obtain clear semantics for small windows, such as 16, 32, etc.
For such a small window, the gain on the SeLo task is not significant due to too much focus on small or partial targets.
While for large windows such as 1024 and 2048, such sliding slices contain too much semantic information.
For RS images with complex scenes, we find that heavy semantic information tends to cause schema collapse of the retrieval model, which produces rather discrete SeLo maps.
Therefore, we try to use intermediate scales to cover most retrieval scenarios when performing this task.
But we need to emphasize that the choice of scale is still an ill-posed problem, which needs to be further fine-tuned on real data.
Such an approach is unavoidable because academia still needs a more consistent setting to be able to study the strengths of the method.
Besides, for the choice of scale, we also conduct further ablation experiments (Table IV, Fig. 9) to explore the retrieval performance at different scales.
}

\textcolor{black}{
On the other hand, different targets in large-scale RS images often have large-scale differences.
In addition to the mitigation of multi-scale clipping, such differences should also be faced by the retrieval model.
Model often requires multi-scale design and corresponding data enhancement to maintain robust adaptation for multi-scale information, which needs to be continuously optimized like other modules.
}

\textcolor{white}{ }

\textbf{\textcolor{black}{Q: Which aspect of the current SeLo framework needs to be optimized?}}

\textcolor{black}{The current SeLo framework is still based on cross-modal retrieval, which is an extremely brute-force computation method.
	In order to achieve more efficient SeLo map generation, the framework can be optimized from the following three aspects:}

\begin{itemize}
	
	\item \textcolor{black}{\textbf{Slice: }
		The current slicing method employs a sliding window, which is a brute-force method in the retrieval stage.
		The number of generated slices determines the number of subsequent cross-modal computations, so this stage is critical to the time of the SeLo map generation.
		If the features irrelevant to the retrieval semantics can be ignored statistically when the sliding window is performed, the calculation time of cross-modal similarity will be greatly reduced.
		In addition, how to perform the scale detection of slices to use slices with the best scale is also an important optimization direction.
	}

	\item \textcolor{black}{\textbf{Cross-modal Similarity Calculation: }
	In natural scenarios, current cross-modal retrieval models are developed towards large models [1][2][3], however such principles may not be applicable for SeLo tasks.
	Although large models can provide well multimodal representations, excessive time consumption must also be considered in this task.    
	Attempting to use lightweight methods such as pruning and distillation to reduce the pre-trained model seems to be a feasible method, and we have also made some attempts [4].
	In addition, the performance of the model does matter.
	Currently, for RS cross-modal tasks, the lack of data has become an important obstacle which restricts the development of this field.
	It is imperative to establish an large and comprehensive remote sensing multi-modal dataset.
	Further, trying to introduce multi-modal pre-training in natural scenes, such as using contrastive learning to improve model representation, are all directions that can be attempted.
	}
	\item \textcolor{black}{\textbf{Post-processing: }
	Current post-processing methods perform operations such as normalization and filtering, which has been found to consume a lot of time at this stage.
	We found that the time spent is extremely large at the median filter stage, and this time is closely related to the image size.
	Alleviating the time occupancy of this stage has become a work that needs to be paid attention to.
	In addition, whether a well post-processing method can be designed to obtain a smoother edge probability is also an optimization direction that can be considered.
	}
	
\end{itemize}

\textcolor{black}{
	We call on more researchers to invest in the research of RS multimodal semantic localization, which is indeed a direction with development and application potential.
	With a large amount of data and pre-trained models, semi-supervised SeLo can even be utilized to unify sub-tasks such as detection and segmentation.
	We hope that more influential work can be produced in this field to advance the field of RS cross-modal retrieval and semantic localization.
}

\textcolor{white}{ }

\textbf{\textcolor{black}{References:}}

\textcolor{black}{[1] Radford, A., Kim, J. W., Hallacy, C., Ramesh, A., Goh, G., Agarwal, S., ... \& Sutskever, I. (2021, July). Learning transferable visual models from natural language supervision. In International Conference on Machine Learning (pp. 8748-8763). PMLR.}

\textcolor{black}{[2] Patashnik, O., Wu, Z., Shechtman, E., Cohen-Or, D., \& Lischinski, D. (2021). Styleclip: Text-driven manipulation of stylegan imagery. In Proceedings of the IEEE/CVF International Conference on Computer Vision (pp. 2085-2094).}

\textcolor{black}{[3] Wang, Z., Yu, J., Yu, A. W., Dai, Z., Tsvetkov, Y., \& Cao, Y. (2021). Simvlm: Simple visual language model pretraining with weak supervision. arXiv preprint arXiv:2108.10904.}

\textcolor{black}{[4] Z. Yuan et al., "A Lightweight Multi-scale Crossmodal Text-Image Retrieval Method In Remote Sensing," in IEEE Transactions on Geoscience and Remote Sensing, doi: 10.1109/TGRS.2021.3124252.}

\end{appendices}
	
\end{document}